\documentclass{article} % For LaTeX2e
\usepackage{iclr2026_conference,times}
\iclrfinalcopy

\setcitestyle{numbers,square}

\usepackage{amsmath,amsfonts,bm}

\def\eqref#1{equation~\ref{#1}}
\def\1{\bm{1}}

\DeclareMathAlphabet{\mathsfit}{\encodingdefault}{\sfdefault}{m}{sl}
\SetMathAlphabet{\mathsfit}{bold}{\encodingdefault}{\sfdefault}{bx}{n}

\usepackage{hyperref}
\usepackage{url}
\usepackage{amssymb}
\usepackage{booktabs}
\usepackage{multirow}
\usepackage{utfsym}
\usepackage{pifont,bbding,makecell}
\usepackage{array}
\usepackage{graphicx}
\usepackage{enumitem}
\usepackage{booktabs}
\usepackage{caption}

\usepackage{tabularx}

\usepackage{graphicx}

\usepackage{subcaption}
\usepackage{xcolor}
\hypersetup{colorlinks=true,linkcolor=red!70!black,linktocpage=false,citebordercolor=blue!70!black,citecolor=blue!70!black,anchorcolor=blue!70!black}

\definecolor{cite_color}{RGB}{190,0,60}
\definecolor{link_color}{RGB}{0,102,102}%  green-style
\definecolor{link_color}{RGB}{153, 0,0}  %  red
\definecolor{url_color}{RGB}{153, 102,  0}
\definecolor{emp_color}{RGB}{0,0,255}
\hypersetup{
 colorlinks,
 citecolor=cite_color,
 linkcolor=link_color,
 urlcolor=url_color}

\title{JTok: On Token Embedding as another Axis of Scaling Law via Joint Token Self-modulation}

\author{
Yebin Yang$^{12}$, Huaijin Wu$^{1}$, Fu Guo$^{2}$, Lin Yao$^{2}$, Xiaohan Qin$^{1}$, \textbf{Jingzhi Wang}$^{1}$, \\
\textbf{Debing Zhang}$^{2}$, \textbf{Junchi Yan}$^{1}$\thanks{Corresponding author: yanjunchi@sjtu.edu.cn.} \\ $^{1}$ School of AI, Shanghai Jiao Tong University \\
$^{2}$ Hi Lab, Xiaohongshu Inc.
}

\begin{document}

\maketitle

\begin{abstract}
LLMs have traditionally scaled along dense dimensions, where performance is coupled with near-linear increases in computational cost. While MoE decouples capacity from compute, it introduces large memory overhead and hardware efficiency challenges. To overcome these, we propose token-indexed parameters as a novel, orthogonal scaling axis that decouple model capacity from FLOPs. Specifically, we introduce \textbf{Joint-Token (JTok)} and \textbf{Mixture of Joint-Token (JTok-M)}, which augment Transformer layers with modulation vectors retrieved from auxiliary embedding tables. These vectors modulate the backbone via lightweight, element-wise operations, incurring negligible FLOPs overhead. 
Extensive experiments on both dense and MoE backbones, spanning from 650M (190M + 460M embedding) to 61B (17B + 44B embedding) total parameters, demonstrate that our approach consistently reduces validation loss and significantly improves downstream task performance (e.g., +4.1 on MMLU, +8.3 on ARC, +8.9 on CEval). Rigorous isoFLOPs analysis further confirms that JTok-M fundamentally shifts the quality–compute Pareto frontier, achieving comparable model quality with 35\% less compute relative to vanilla MoE architectures, and we validate that token-indexed parameters exhibit a predictable power-law scaling behavior.
Moreover, our efficient implementation ensures that the overhead introduced by JTok and JTok-M remains marginal.

\end{abstract}

% \vspace{-7pt}
\section{Introduction}
% \vspace{-7pt}

The development of large language models (LLMs) is closely linked to the scaling laws of Transformer architectures~\cite{vaswani2017attention}. The conventional approach to enhancing model performance involves increasing the number of parameters and training tokens, which typically yields smooth power-law improvements~\citep{kaplan2020scaling, hoffmann2022training}. However, both FLOPs and GPU memory requirements scale approximately linearly with model size, while the availability of high-quality text data is becoming increasingly limited. As a result, simply scaling dense models leads to diminishing marginal returns and may even cause performance degradation in data-constrained settings~\cite{willwerunoutofdata, kim2025pre}.

To address the inefficiencies of dense scaling, Mixture-of-Experts (MoE) architecture~\cite{fedus2022switch,deepseekmoe} has emerged as a promising alternative, decoupling model capacity from computation by leveraging sparsely activated expert subnetworks while maintaining approximately constant active FLOPs. However, since the relationship between loss and sparsity follows a log-linear trend~\cite{tian2025towards}, the benefits of expert sparsity also saturate rapidly. Moreover, sparse models also exhibit lower sample efficiency, requiring larger datasets to reach convergence~\cite{krajewski2024scaling,krajewski2025scaling}, alongside significant engineering challenges to ensure hardware efficiency and low latency~\cite{huang2024toward, deepseekv3} for routing balance~\cite{kim2024scaling}.

Recognizing that scaling dense parameters, data, and expert sparsity all encounter fundamental bottlenecks, in this paper, we explore token-indexed parameters as an orthogonal and complementary scaling dimension. Specifically, we propose a module called Joint-Token (JTok), which augments each Transformer layer by applying a token-specific modulation vector—retrieved from learned embedding tables—to gate the MLP residual~\cite{residual} via lightweight Hadamard products. Furthermore, we extend JTok into a sparse variant, Mixture of JTok (JTok-M), which maintains a pool of modulators per token and employs a lightweight router to select context-appropriate mixtures.

This architectural design explicitly addresses the bottlenecks inherent in the aforementioned scaling paradigms. First, by relying on element-wise modulation instead of dense matrix multiplications, JTok injects substantial capacity with minimal FLOPs overhead, thereby overcoming the efficiency limits of dense scaling. Second, unlike standard MoE models that incur high communication and HBM overhead, our retrieval-based mechanism is largely decoupled from the backbone computation. This enables asynchronous prefetching of parameters, which can be overlapped with backbone execution—effectively mitigating the latency typically associated with high sparsity.

Extensive experiments reveal that JTok-M fundamentally shifts the quality-compute Pareto frontier, delivering consistent performance gains even as the backbone model scale increases. Furthermore, we explicitly validate the scalability of this dimension: the loss exhibits a log-linear trend with the number of token-indexed parameters, confirming that this axis follows a scaling law similar to dense parameters. \textbf{To summarize, our contributions are as follows:}

\begin{itemize}[leftmargin=*, itemsep=0pt, topsep=0pt]
\item We introduce token-indexed parameters as a novel scaling axis that expands model capacity without increasing FLOPs, offering a complementary axis to traditional dense and sparse scaling.
\item We propose JTok and JTok-M, practical architectures that leverage token-indexed modulation to enhance both dense and MoE models with negligible FLOPs overhead.
\item We show that JTok-M consistently accelerates convergence and reduces training loss, directly translating into significant downstream performance gains. Notably, on a large-scale 17B MoE backbone, our approach delivers substantial improvements (e.g., +4.1 on MMLU, +8.3 on ARC, and +8.9 on CEval), validating its robustness in enhancing both data efficiency and model generalization.
\item We establish token-indexed parameters as a highly efficient and scalable dimension. Rigorous iso-compute analysis confirms a 35\% compute saving that remains stable across backbone scales , while the token-indexed parameters themselves exhibit a predictable log-linear scaling behavior, offering a proven trajectory for future model expansion.
\item We propose efficient implementation strategies with which, training throughput loss is less than 7\%; for inference, there is no additional GPU memory footprint, with a moderate latency increase of $\le$ 7.3\%.
\end{itemize}

\begin{figure*}[t]
  \centering
  \vspace{-10pt}
  \includegraphics[width=0.9\textwidth]{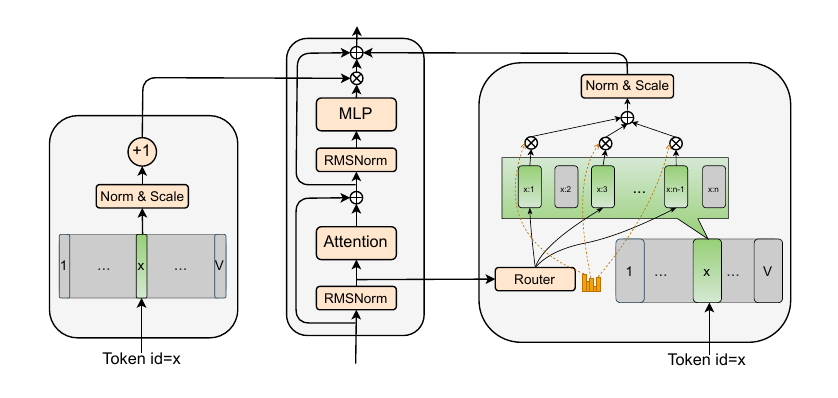}
  % \vspace{-10pt}
  \caption{\textbf{Architecture of JTok/JTok-M.} JTok (left) augments each Transformer layer with a token-indexed table. Each token retrieves a modulation vector, applies norm and a learnable per-dimension scaling, and forms a lightweight multiplicative gate to modulate the FFN update via element-wise products. JTok-M (right) generalizes JTok by maintaining a pool of token-indexed modulators and using a router conditioned on the hidden state to select a sparse Top-$K$ mixture per token; the mixed modulator is normalized+scaled and injected as an additional residual alongside the backbone update. Both are plug-in bypass modules implemented with table lookups and element-wise operations, allowing retrieval to be overlapped with backbone and adding negligible compute.}
  \vspace{-10pt}
  \label{fig:model_arch}
\end{figure*}

% \vspace{-7pt}
\section{Related Works} \label{related_work}
% \vspace{-7pt}

\textbf{Scaling Law.} 
The Kaplan scaling law~\cite{kaplan2020scaling} empirically shows that LLM performance follows power-law with parameters, data, and compute, enabling extrapolation across orders of magnitude. Chinchilla~\cite{hoffmann2022training} refined these findings into compute-optimal training prescriptions, arguing that many earlier LLMs were under-trained and that optimal scaling under fixed training compute requires increasing training tokens roughly proportionally with model parameters. \cite{sardana2401beyond} incorporate deployment considerations by explicitly accounting for inference cost in the scaling objective, showing that when inference demand is very high, the overall-optimal strategy may favor smaller models trained for longer to minimize end-to-end compute. 

In parallel, scaling-law frameworks have been extended to Mixture-of-Experts (MoE) architectures. \cite{krajewski2024scaling} introduced a granularity hyperparameter for MoE models and found MoE consistently outperforms compute-matched dense models, with gains increasing at larger scales. \cite{tian2025towards} proposed the efficiency leverage metric and empirically linked efficiency gains to factors such as expert-activation fraction via power laws, predicting that well-configured MoE can match dense performance with substantially less compute. Overall, these works systematize LLM scaling laws across compute allocation, inference cost, and alternative architectures, offering practical guidance for improving efficiency.

\textbf{Vocabulary Scaling}
emerges as another factor for quality and efficiency.~\cite{tao2024scaling} provides a compute-optimal rule indicating that larger models benefit from larger vocabularies and show fixed-FLOPs gains. In continual-training scenarios,~\cite{takase2024large} replacing the old vocabulary with a better-matched one outperforms keeping the original tokenizer.~\cite{huang2025over} decouples input/output vocabularies to enlarge only the input side with no extra inference cost;
% ~\cite{deiseroth2024t} goes tokenizer-free via sparse character n-grams to cut embedding/LM-head size;
SuperBPE~\cite{liu2025superbpe} extends BPE with a simple pretokization curriculum—learn subwords first, then allowing merges across whitespace to form multi-word tokens—improving encoding efficiency and downstream performance.
BLT~\cite{pagnoni2025byte} replaces fixed-vocabulary tokens with dynamically sized byte patches as the main computation units, enabling tokenizer-free scaling with improved efficiency and robustness.

\textit{Distinction from our approach:} While existing methods primarily scale the vocabulary dimension V—often utilizing techniques like hash N-grams~\citep{huang2025over,yu2025scaling}—they are inherently constrained by combinatorial limits. Such expansions tend to capture fixed, local surface patterns without deepening semantic understanding. In contrast, we scale along the feature dimension $d$, providing a high-dimensional, context-interactive space in which tokens can acquire richer semantics through attention-mediated interactions during training.

\textbf{Mixture of Experts.}
MoE architectures have advanced toward sparser routing, finer-grained experts, and greater deployability. The sparsely-gated MoE of ~\cite{shazeer2017outrageously} activates only a small subset of FFN experts per example via a trainable gate with an auxiliary load-balance loss, scaling capacity to tens of billions of parameters with limited extra compute and validating conditional computation for language modeling and translation. Switch Transformer~\cite{fedus2022switch} further simplifies routing by selecting a single expert per token, yielding a favorable speed-quality trade-off under fixed compute and enabling trillion-parameter scaling.~\cite{deepseekmoe, deepseekv2} pushes extreme expert specialization by partitioning experts more finely and isolating shared experts for generic knowledge, improving utilization and reducing redundancy, and reporting competitive or superior performance to larger dense/MoE baselines at comparable cost. To ease deployment and accelerate inference,~\cite{jie2025mixture} reparameterizes trained FFN experts into lookup tables for inference, retrieving precomputed expert outputs on demand to slash VRAM usage and avoid real-time expert compute, which yields dense-like latency in their setting.

\textbf{Large Memory Layer.}
Large memory layers expand model capacity by decoupling parameters from compute. Product Key Memory~\citep{lample2019large} inserts a very large, sparsely accessed key-value table into the FFN, adding minimal extra FLOPs while improving language modeling at scale. PEER~\cite{he2024mixture} further extends product-key style routing to sparsely select from a pool of over a million tiny experts, improving the performance-compute trade-off in a fine-grained MoE-like regime. Ultra-Sparse Memory~\citep{huang2024ultra} introduces an ultra-sparse memory layer that activates only a few memory slots per token and reports reduced inference latency while maintaining performance. Memory Layers at Scale~\citep{berges2024memory} shows that replacing some FFNs with trainable key-value memory layers scales to very large memory capacity and can outperform dense models using substantially more FLOPs, while also being competitive with MoE under matched budgets.

% \vspace{-7pt}
\section{Methodology}
% \vspace{-7pt}
\label{method}

\subsection{Preliminary}

\subsubsection{Pre-Norm Transformer Block}
\label{sec:prenorm}

We adopt the Pre-Norm Transformer~\cite{prenorm} as our backbone, a prevalent architecture in current SOTA open-source models~\cite{qwen3next, deepseekv3} due to its robustness in stabilizing training dynamics and gradients~\cite{zhu2024hyper}. In this setting, each sub-module operates on a normalized hidden state and contributes an additive residual update. Concretely, for token $x$ at layer $\ell$,
\begin{align}
\Delta \mathbf{a}_x^{\ell} &= \mathrm{Attn}^{\ell}\!\left(\mathrm{RMSNorm}(\mathbf{h}_x^{\ell})\right), \\
\tilde{\mathbf{h}}_x^{\ell} &= \mathbf{h}_x^{\ell} + \Delta \mathbf{a}_x^{\ell}, \\
\Delta \mathbf{m}_x^{\ell} &= \mathrm{FFN}^{\ell}\!\left(\mathrm{RMSNorm}(\tilde{\mathbf{h}}_x^{\ell})\right), \\
\mathbf{h}_x^{\ell+1} &= \tilde{\mathbf{h}}_x^{\ell} + \Delta \mathbf{m}_x^{\ell},
\end{align}
where $\mathrm{FFN}^{\ell}(\cdot)$ is the MLP for dense model and becomes sparsely-activated experts for MoE.

\subsubsection{Scaling laws and isoFLOP profiles}
\textbf{Scaling Law Form.}
Neural scaling laws provide an empirical yet principled lens through which to reason about the quality-compute trade-off of LLM training.
According to the Kaplan scaling law~\cite{kaplan2020scaling}, the test loss can be modeled as:
\begin{equation}
\label{eq:kaplan}
\mathcal{L}(N_c,D)
=
\Biggl[
\Bigl(\frac{A}{N_c}\Bigr)^{\frac{\alpha}{\beta}}
+
\frac{B}{D}
\Biggr]^{\beta},
\end{equation}
where $N_c$ denotes the compute-intensive parameters (in MoE cases, it refers to activated parameters), excluding both embedding and LM prediction head; $D$ is the number of training tokens, and $A,B,\alpha,\beta$ are empirical constants.

\textbf{IsoFLOPs Profiles.} \label{iso-protocal}
IsoFLOPs profiles provide an empirical procedure for estimating the compute-optimal allocation between model size and data under a fixed training FLOPs budget~\citep{hoffmann2022training}.
Under standard Transformer FLOPs accounting, training a model with $N_c$ compute-intensive parameters\footnote{Here, the parameters consist of attention qkvo matrices and mlp parameters (or activated experts for MoE).} 
for $D$ tokens costs approximately $C \approx 6N_cD$ FLOPs~\cite{pearce2024reconciling}.
The compute-optimal loss at budget $C$ is defined as
\begin{equation} \label{eq:L_star}
    \mathcal{L}^{\ast}(C) \;=\;  \mathcal{L}(N_c, D), \quad s.t. \; 6N_cD = C
\end{equation}
To estimate $L^{\ast}(C)$ in practice, a set of target budgets $\{C_i\}$ is fixed. For each $C_i$, model size $N_c$ is swept over a grid, and the corresponding token budget is set to $D = C_i/(6N_c)$ so that each run matches the same total FLOPs. After training, the final held-out loss is recorded for each configuration, yielding an IsoFLOPs curve of loss versus $N_c$ at constant $C_i$, which typically exhibits a U-shaped profile with a clear minimum. A quadratic fit around the valley can be used to estimate the loss-optimal model size $N_c^{\ast}(C_i)$ and token count $D^{\ast}(C_i)$ for each budget. Collecting minima across budgets forms the empirical efficient frontier $\{(C_i, L^{\ast}(C_i))\}$.

% \vspace{-7pt}
\subsection{JTok} 
\label{JTok}
% \vspace{-7pt}

JTok scales the token embedding along hidden dimension $d$.
As illustrated in Fig.~\ref{fig:model_arch} (left), every transformer layer $\ell$ keeps a learnable table $\mathbf{E}^\ell \in \mathbb{R}^{V\times d}$. Each token retrieves a vector $\mathbf{E}^\ell[x]$ with its ID $x \in [V]$ and gates the backbone module's residual updates via Hadamard products.

\textbf{Mechanism.} Let $x$ be the token id, 
% $\mathbf{h}^\ell _x \in \mathbb{R}^d$ be the input hidden state of token $x_t$ in layer $\ell$, 
$\Delta \mathbf{m}^\ell _x$ be the MLP module increments. JTok forms multiplicative gate $\mathbf{p}_{x}^{\ell} \in \mathbb{R}^d$:
\begin{equation}
\label{eq:tokBinder_vector}
   \mathbf{p}_{x}^{\ell} = \mathbf{1} + \mathbf{s}^\ell \odot \operatorname{Norm_{\varepsilon}}(\mathbf{E}^\ell[x]),
\end{equation} 
where $\mathbf{s}^\ell \in \mathbb{R}^d$ is a learnable per-dimension scaler and $\mathrm{Norm}_\varepsilon(\mathbf{u})=\frac{\mathbf{u}}{\lVert \mathbf{u}\rVert_2+\varepsilon}$ ($\varepsilon$ is a small constant to avoid division by zero). We empirically validate the effectiveness of this normalization term in Appendix~\ref{app:ablation_norm}. The gated MLP increments is
\begin{equation}
   \Delta \hat{\mathbf m}_x ^\ell = \Delta \mathbf{m}_x ^\ell \odot \mathbf{p}_{x}^{\ell},
\end{equation}
which then adds to the backbone residual:
\begin{equation} \label{eq:JTok_update}
   \mathbf{h}_x^{\ell+1} =\tilde{\mathbf{h}}_x^{\ell} + \Delta \hat{\mathbf m}_x ^\ell.
\end{equation}

% \vspace{-7pt}
\subsection{Mixture of JTok (JTok-M)} \label{JTok-M}
% \vspace{-7pt}

To further unlock the potential of token-indexed parameters, we introduce JTok-M. JTok-M generalizes from a static mapping to a dynamic, context-aware routing framework. By maintaining a larger pool of modulators and leveraging the hidden state as a contextual signal to guide the selection, JTok-M adaptively selects a sparse mixture of parameters for each token instance. This design enables the model to capture context-dependent semantics and significantly expands the parameter space while preserving the efficiency of sparse retrieval. As shown in Fig.~\ref{fig:model_arch} (right).

\textbf{Mechanism.}
Each token is equipped with a pool of $n_e$ modulators per layer and uses a router to pick top-$K$ of them given the hidden state.

Formally, for token $x$, let $\mathrm{RMSNorm}(\mathbf{h}_x^\ell)$ be the input of attention in layer $\ell$. Each layer maintains a pool
$\{\mathbf{E}_i^\ell \in \mathbb{R}^{V\times d}\}_{i=1}^{n_e}$.
A linear router computes logits
\begin{equation}
   \mathbf{g}^\ell_x=(\mathrm{RMSNorm}(\mathbf{h}_x^\ell))^\top \mathbf{R}^\ell \;\in\; \mathbb{R}^{n_e},~\mathbf{R}^\ell \in \mathbb{R}^{d\times n_e},
\end{equation}
selects $G^\ell_x=\mathrm{TopK}(\mathbf{g}^\ell_x, K)$, and forms normalized weights
$w_i^\ell=\frac{\sigma(g_i^\ell)}{\sum_{j\in G^\ell_x}\sigma(g_j^\ell)}$ for $i\in G^\ell_x$ with a Sigmoid $\sigma$~\citep{sigmoidgating}.
The mixed token-indexed vector is:
\begin{equation}
   \mathbf{e}^\ell _x \;=\; \sum_{i\in G^\ell_x} w_i^\ell \, \mathbf{E}_i^\ell[x]\in\mathbb{R}^d.
\end{equation}
We normalize and apply a learnable element-wise scaler $\mathbf{s}_{\mathrm{JTok-M}}^{\ell}\in\mathbb{R}^d$, producing the additive residual injection. To ensure the variance of the hidden states controllable~\citep{radford2019language}, we scale with an extra factor $\frac {1}{\sqrt{2 N_l}}$, where $N_l$ is the number of layers of the backbone. The ablation of this scaling factor is provided in Appendix~\ref{app:ablation_downscaling}.
\begin{equation}\label{eq_down_scale}
   \Delta \mathbf{r}_{x}^\ell \;=\; \frac {1}{\sqrt{2 N_l}} \cdot  \mathbf{s}_\mathrm{JTok-M}^{\ell} \odot \mathrm{Norm}_\varepsilon\!\big(\mathbf{e}^\ell _x \big),
\end{equation} 
Finally, $\Delta \mathbf{r}_{x}^\ell$ is fused into the layer write-back together with MLP output $\Delta \mathbf{m}_x^\ell$:
\begin{equation} \label{eq:JTok-M_update}
   \mathbf{h}_{x}^{\ell+1} = \tilde{\mathbf{h}}_{x}^{\ell} \;+\; \Delta \mathbf{m}_x^\ell \;+\; \Delta \mathbf{r}_{x}^\ell
\end{equation} 
% \textbf{Training with Load Balancing.} 
To encourage all the embedding experts to be adequately utilized and trained, we incorporate an auxiliary load-balancing loss during training, similar to standard practice in MoE models. Implementation details are provided in Appendix~\ref{app:loadbalance}.

% \vspace{-7pt}
\subsection{System Efficiency}
\label{sec:efficiency}
% \vspace{-7pt}

Token-indexed parameters shift the main system concern from GEMMs to parameter access and memory footprint.
While the Transformer backbone is typically compute-bound, token-indexed modules have negligible FLOPs but are dominated by memory access. Accordingly, our system design targets:

1. keeping token-indexed access off the GEMM critical path by overlapping retrieval with backbone compute via asynchronous prefetch/overlap (Sec.~\ref{eff:overlap}), and minimizing memory traffic via token deduplication (Sec.~\ref{eff:zips}); \\
2. reducing the HBM footprint of token-indexed tables via embedding model parallelism (Sec.~\ref{eff:embp}) and CPU offloading (Sec.~\ref{eff:cpuoffload}).

These strategies allow JTok/JTok-M to scale to massive memory capacities with minimal impact on efficiency.

% \vspace{-7pt}
\subsubsection{Computational Complexity Analysis}
% \vspace{-7pt}
Per token and per layer, JTok adds a normalization, a Hadamard product for gating, and a residual write-back, all of which are $\mathcal{O}(d)$.\footnote{From the roofline~\cite{roofline} perspective, these element-wise Ops could be limited by memory traffic and kernel-launch. Kernel fusion can amortize launches and avoid extra memory round-trips.}

JTok-M additionally computes router logits and mixes top-$K$ modulators.
Routing is a single matrix-multiply with cost $\mathcal{O}(dn_e)$, and mixture application is $\mathcal{O}(Kd)$.
In practice $n_e$ and $K$ are small constants, so the overall overhead remains linear in $d$ and is negligible
compared to the backbone's $\Theta(d^2)$ attention/FFN computes.

% \vspace{-7pt}
\subsubsection{Memory Access and Zipfian Patterns} \label{eff:zips}
% \vspace{-7pt}

The dominant overhead of JTok/JTok-M is memory traffic from table lookups.
For each layer, JTok reads $d$ elements per token, while JTok-M reads $Kd$. However, token frequencies follow a Zipfian distribution~\cite{zipf}, so high-frequency tokens are repeatedly accessed.
This enables \emph{token deduplication}: unique token ids in a micro-batch are gathered once and
then scattered back to all occurrences, avoiding redundant reads. 
% Implementation details are deferred to Appendix~\ref{app:system_dedup}.

% \vspace{-7pt}
\subsubsection{Decoupling and Asynchronous Overlap} \label{eff:overlap}

JTok/JTok-M are bypass modules, decoupled from backbone, so their execution can be
scheduled independently of attention/FFN computation, enabling overlapping memory access with compute.
Concretely, embedding gathers for some layer can be issued asynchronously while
the backbone executes GEMMs for the current layer, and the retrieved vectors are fused only at the
layer write-back (Eqs.~\eqref{eq:JTok_update} and~\eqref{eq:JTok-M_update}).
As a result, a substantial portion of the retrieval latency can be hidden under compute, making the end-to-end
throughput impact small in practice.

% \vspace{-7pt}
\subsubsection{Embed-Parallel in Training} \label{eff:embp}
% \vspace{-7pt}

Token-indexed tables are large but sparsely accessed, making them well-suited for \emph{embedding model parallelism}.
Tables can be sharded across GPUs, reducing per-device memory footprint, allowing larger micro-batches and increasing training throughput.
For JTok, the cross-device communication per token is $d$ elements.
For JTok-M, a naive implementation would communicate $Kd$ elements, but this can be avoided by
\emph{owner-rank premixing}: the device that owns the selected modulators performs the weighted sum locally,
and only the mixed vector ($d$-sized) is communicated.

% \vspace{-7pt}
\subsubsection{CPU Offloading in Inference} \label{eff:cpuoffload}
% \vspace{-7pt}

Since token-indexed parameters are accessed sparsely, the host-to-device transfer volume depends on the number of
requested vectors, not on the full table size.
Per layer, transferring the retrieved values scales as $\mathcal{O}(d)$ for JTok and $\mathcal{O}(Kd)$ for JTok-M,
independent of $V$ and of the overall parameter capacity stored on the host.
This property makes CPU offloading a practical option during inference to save HBM usage. Specifically, the tables can reside in CPU memory, while only the small set of vectors needed by the current batch are streamed
to GPUs and overlapped with backbone execution.
% Appendix~\ref{app:system_offload} provides implementation details.

% \vspace{-7pt}
\subsection{Scaling Hypothesis of Token-Indexed Parameters}
% \vspace{-7pt}
\label{method_scaling}

Building on the Kaplan scaling law (Eq.~\ref{eq:kaplan}) and the FLOPs constraint $C\approx 6N_cD$ introduced in the Preliminary,
we introduce one core assumption: effective parameter count $N_{\mathrm{eff}}$.
This assumption integrates token-indexed parameters into the scaling law and yields a \emph{scale-invariant} isoperformance compute saving in the compute-optimal regime.

\textbf{Core assumption: effective parameters $N_{\mathrm{eff}}$.}
We denote the backbone compute-intensive parameters as $N_c$ and the token-indexed parameters as $N_n$.
Let $\eta\triangleq N_n/N_c$ be the parameter expansion ratio.
JTok-M retrieves and injects token-indexed vectors per token and per layer; its usable capacity depends on routing sparsity.
We capture this with a JTok-M architecture-dependent discount function $\gamma(\rho)$, where $\rho$ denotes the JTok-M activation sparsity ($\rho=K/n_e$).
We define
\begin{equation} \label{eq:neff}
N_{\mathrm{eff}}\;\triangleq\;N_c+\gamma(\rho)N_n
\;=\;N_c\bigl(1+\eta\gamma(\rho)\bigr).
\end{equation}
Importantly, $N_{\mathrm{eff}}$ characterizes \emph{effective capacity}, whereas the dominant training FLOPs are still governed by $N_c$, $D$.

\textbf{Incorporating $N_{\mathrm{eff}}$ into the Kaplan scaling law.}
Starting from Eq.~\ref{eq:kaplan}, we replace the $N_c$ with $N_{\mathrm{eff}}$:
\begin{equation}\label{eq:JTok-M_scaling}
\mathcal{L}_{\mathrm{JTok-M}}(N_c,D;\eta,\rho)
% &=\Biggl[\Bigl(\frac{A}{N_{\mathrm{eff}}}\Bigr)^{\frac{\alpha}{\beta}}+\frac{B}{D}\Biggr]^{\beta}\\
=\Biggl[\Bigl(\frac{A_{\mathrm{JTok-M}}}{N_c}\Bigr)^{\frac{\alpha}{\beta}}+\frac{B}{D}\Biggr]^{\beta},
\end{equation}
where $A_{\mathrm{JTok-M}}\triangleq A/(1+\eta\gamma(\rho))$.
This highlights that JTok-M is equivalent to rescaling the constant in the model-size term, while leaving the data term $B/D$ and the dominant FLOPs form unchanged.

\textbf{Isoperformance compute saving under compute-optimal training.}
Let $\mathcal{L}^*(C)$ denote the compute-optimal efficient frontier of backbone model defined in Eq.~\ref{eq:L_star}.
Similarly, the JTok-M compute-optimal frontier becomes a \emph{multiplicative downward shift}:
\begin{equation}
\label{eq:frontier_JTok-M}
\mathcal{L}_{\mathrm{JTok-M}}^*(C;\eta,\rho)
=\bigl(1+\eta\gamma(\rho)\bigr)^{-\frac{\alpha\beta}{\alpha+\beta}} \cdot\mathcal{L}^*(C).
\end{equation}
Together, for any target test loss $\mathcal{L}^{\star}$, the minimal compute required satisfies
\begin{equation}
\label{eq:compute_saving}
C_{\mathrm{JTok-M}}^{\star}(\mathcal{L}^{\star})
=\frac{1}{1+\eta\gamma(\rho)}\,C_{\mathrm{base}}^{\star}(\mathcal{L}^{\star}).
\end{equation}
The key insight of Eq.~\ref{eq:compute_saving} is that the \textbf{compute saving ratio is independent of the backbone scale} and depends only on JTok-M architectural hyperparameters (e.g., $\eta$, $\rho$, and $\gamma(\cdot)$). The detailed derivations of Eq.~\ref{eq:frontier_JTok-M} and Eq.~\ref{eq:compute_saving} are provided in Appendix~\ref{app:derivation}.

% \vspace{-7pt}
\section{Experiments}
\label{exps}

\subsection{Main Quality Results}
\label{sec:exp:main_quality}

\subsubsection{Experimental Setup}

We first validate that token-indexed parameters consistently improve quality across diverse architectures (including both dense and MoE models) and backbone scales (up to 17B), before examining scaling behaviors and efficiency.

\begin{figure*}[t]
   \centering
   \begin{subfigure}[b]{0.32\textwidth}
      \includegraphics[width=\textwidth]{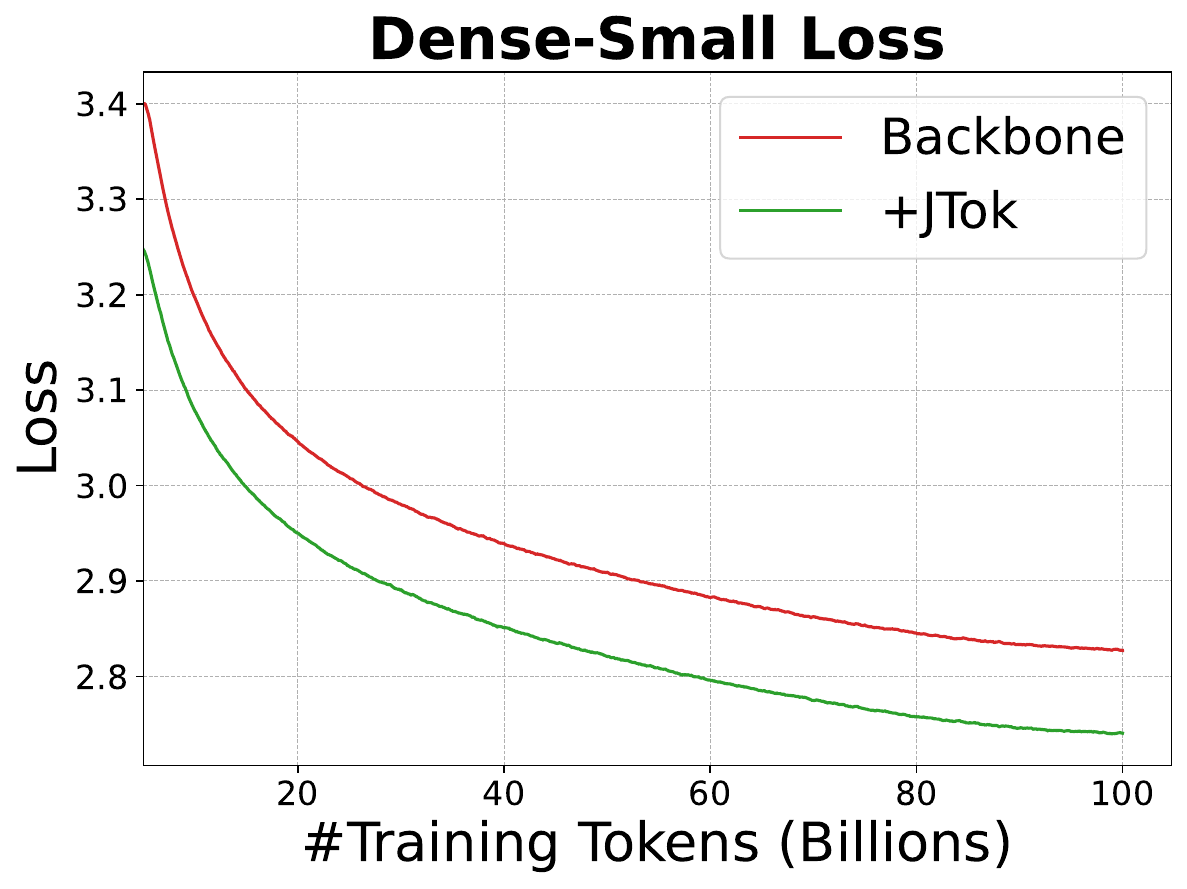}
      % \caption{Dense-S (190M)}
      \label{fig:dense_s} 
   \end{subfigure}
   \hfill
   \begin{subfigure}[b]{0.32\textwidth}
      \includegraphics[width=\textwidth]{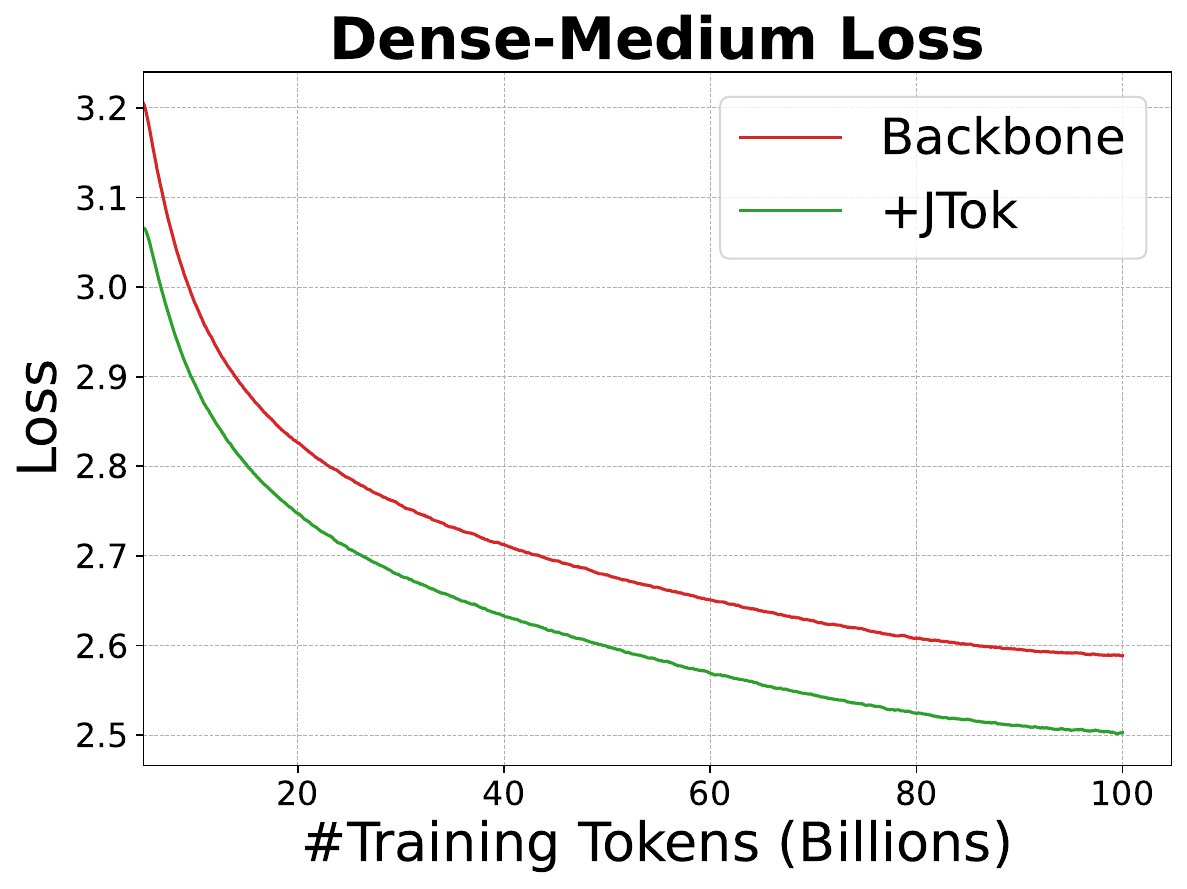}
      % \caption{Dense-M (505M)}
      \label{fig:dense_m}
   \end{subfigure}
   \hfill
   \begin{subfigure}[b]{0.32\textwidth}
      \includegraphics[width=\textwidth]{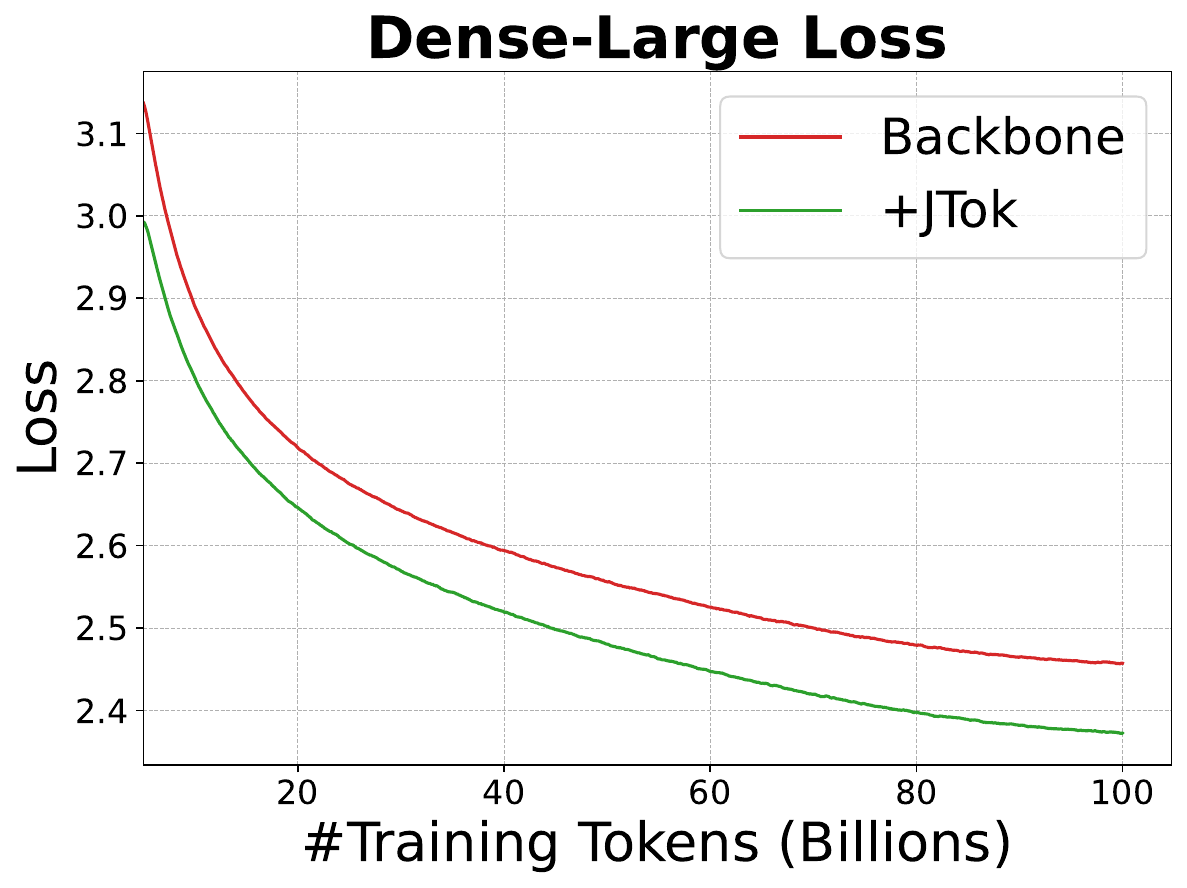}
      % \caption{Dense-L (1B)}
      \label{fig:dense_l}
   \end{subfigure}
   
  %  \vspace{5pt}
   
   \begin{subfigure}[b]{0.32\textwidth}
      \includegraphics[width=\textwidth]{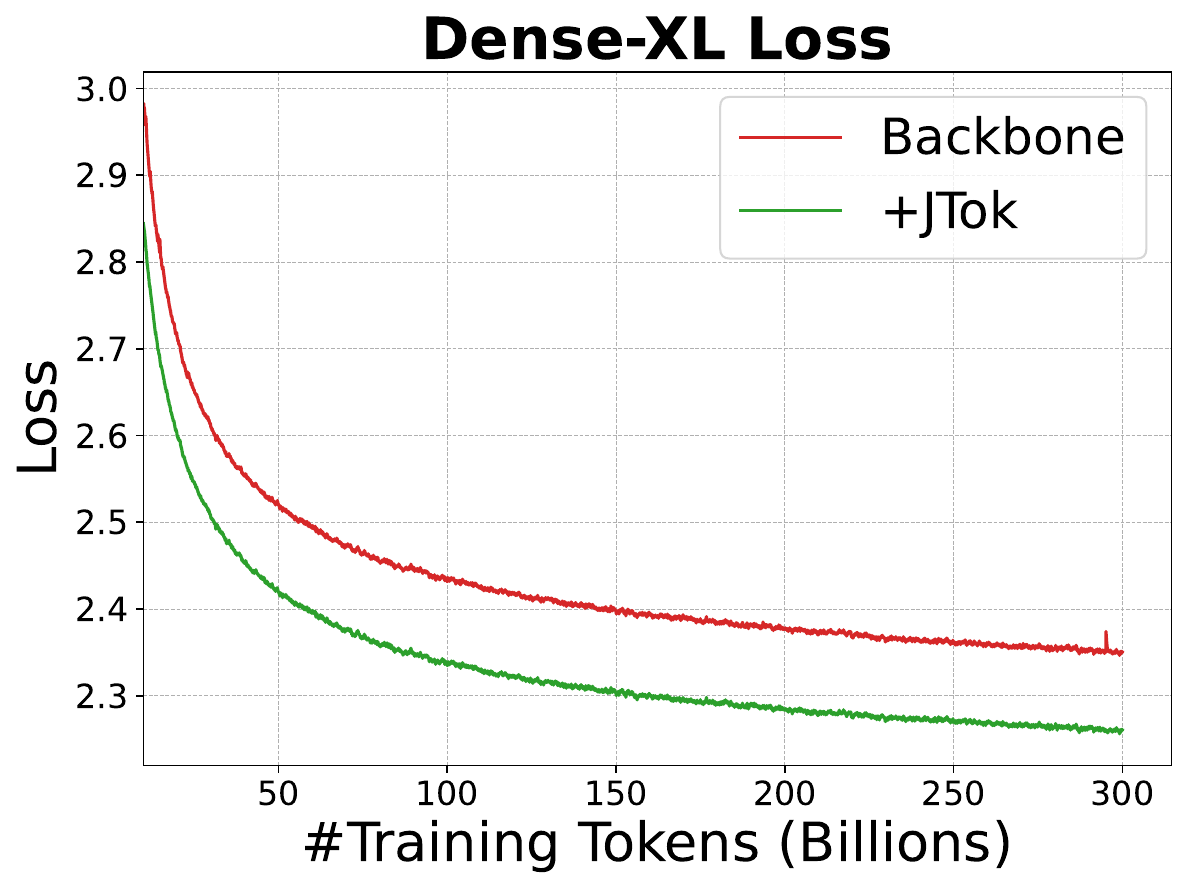}
      % \caption{Dense-XL (1.5B)}
      \label{fig:dense_xl}
   \end{subfigure}
   \hfill
   \begin{subfigure}[b]{0.32\textwidth}
      \includegraphics[width=\textwidth]{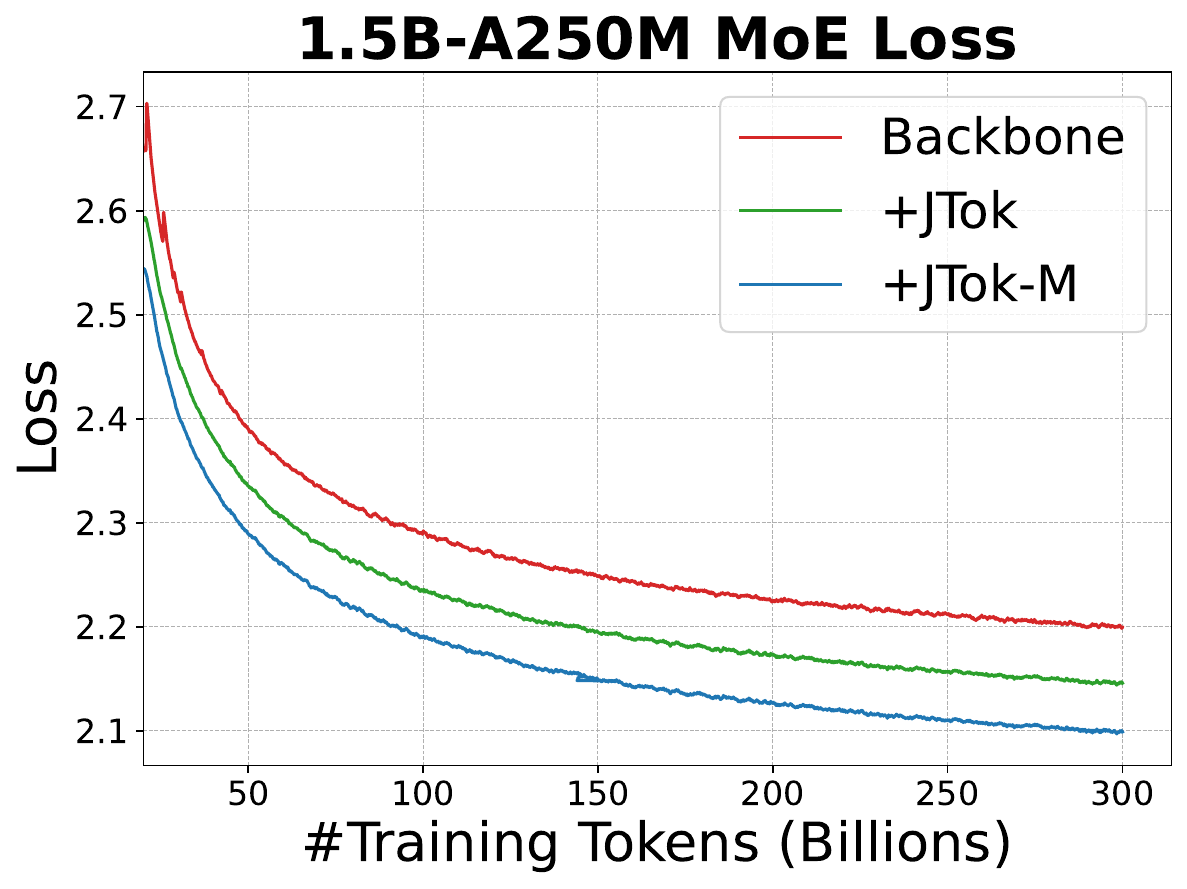}
      % \caption{MoE (1.5B-A250m)}
      \label{fig:moe_1_3b}
   \end{subfigure}
   \hfill
   \begin{subfigure}[b]{0.32\textwidth}
      \includegraphics[width=\textwidth]{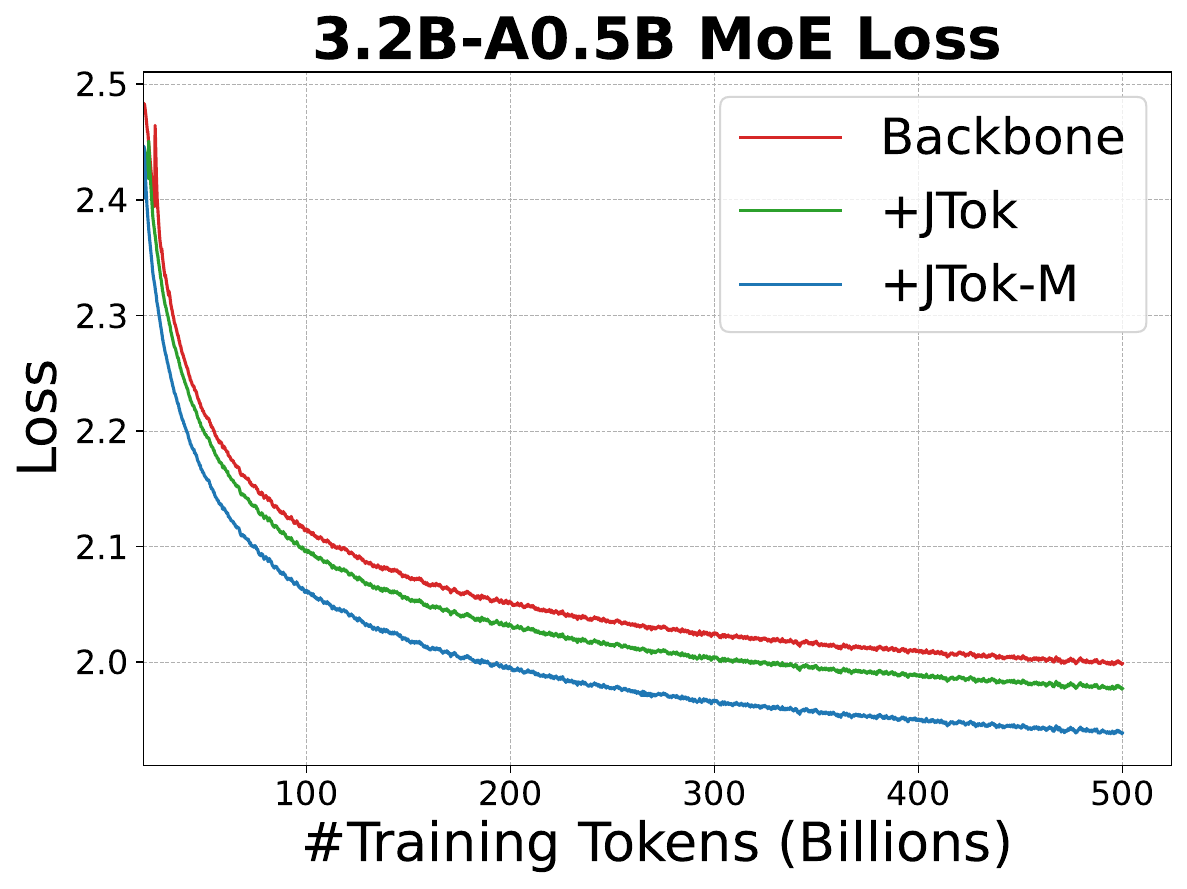}
      % \caption{MoE (3.2B-A0.5B)}
      \label{fig:moe_3b}
   \end{subfigure}

   % \vspace{-7pt}
   \caption{\textbf{Training loss for dense and MoE backbones and corresponding JTok and JTok-M variants.} The top row shows JTok's performance on Dense-S(190M), M(505M), and L(1B) backbones. The bottom row shows results on Dense-XL (1.5B), 1.5B-A250M MoE, and 3.2B-A0.5B MoE backbones. In all settings, JTok (and JTok-M for MoE backbone) achieves a consistently and significantly lower training loss.}
   \label{fig:all_loss_curves}
   % \vspace{-7pt}
\end{figure*}

\textbf{Backbone Models.} We establish backbone using both dense and MoE architectures. For dense models, we use four Qwen-style~\citep{qwen3} models of varying sizes: 190M(S), 0.5B(M), 1B(L), and 1.5B(XL) parameters. 
For MoE models, we use two highly-sparse configurations: one with 250M activated parameters and 1.5B total parameters (1.5B-A250M), and another with 0.5B activated parameters and 3.2B total parameters (3.2B-A0.5B). Each MoE layer contains 145 experts, with one shared expert~\cite{deepseekmoe} and 144 routed experts, from which the top-8 are activated per token. To further investigate the scalability of our approach on large-scale foundation models, we introduce a 17B MoE backbone (17B-A2B). This model features 28 layers with 65 experts (1 shared expert) per layer, activating top-6 per token, resulting in 1.9B activated parameters. For more details about the model configurations, please refer to Appendix~\ref{app:hyper}.

\textbf{JTok/JTok-M Configuration.} We evaluate two methods: JTok and JTok-M. Both are implemented as attached modules that augment the backbone model without modifying its architecture. JTok is benchmarked on the dense-XL, the 1.5B-A250M MoE and the 3.2B-A0.5B MoE backbones. JTok-M is evaluated on the 1.5B-A250M, the 3.2B-A0.5B and the 17B-A2B MoE backbones. For JTok-M, we set the number of modulator experts $n_e=5$ and activate the top $K=2$. See Appendix~\ref{app:hyper} for detail of the additional parameter counts introduced by the JTok and JTok-M modules.

\textbf{Datasets and Training.} All models are pretrained with Megatron-LM framework~\cite{megatron-lm}. The S, M, and L dense models are pretrained for 100B tokens on the Fineweb-edu~\cite{fineweb-edu}, an open-source collection of text dataset. The XL dense model and all MoE models are trained on another high-quality dataset curated from online corpora, which includes general text, code, math, and multilingual content after rigorous filtering. Both XL dense and 1.5B-A250M MoE are trained for 300B tokens, and the 3.2B-A0.5B MoE for 500B tokens. 
The 17B MoE backbone and its JTok-M variant are trained for 570B tokens on the same high-quality curated dataset.
All training configurations for the JTok and JTok-M models are kept identical to those of their corresponding backbones. For detailed training hyperparameters, see Appendix~\ref{app:hyper}.

\textbf{Evaluation.} We evaluate downstream benchmark performances of dense-XL, 1.5B-A250M and 3.2B-A0.5B MoEs as well as corresponding JTok/JTok-M variants using OpenCompass~\cite{opencompass} framework. The downstream benchmarks distribute in 4 domains: \textit{Knowledge} (MMLU~\cite{mmlu}, TriviaQA~\citep{triviaqa}, ARC~\citep{arcC}, GPQA~\citep{gpqa}), \textit{Reasoning} (Hellaswag~\citep{hellaswag}, C3~\citep{c3}, BBH~\citep{bbh}, SocialIQA~\citep{socialiqa}), \textit{Code} (MBPP~\citep{mbpp}, HumanEval~\citep{humaneval}, LiveCodeBench~\citep{livecodebench}) and \textit{Math} (MATH~\citep{math}, GSM8K~\citep{gsm8k}, DROP~\citep{dua2019drop}).

Additionally, for the 17B-A2B MoE and its Jtok-M variant (17B + 44B embedding), we track the accuracy trajectories against the number of training tokens on six representative benchmarks: MMLU, ARC, Hellaswag, and CMMLU~\cite{cmmlu}, CEval~\cite{ceval}, Xiezhi~\cite{xiezhi} to analyze the convergence behavior and performance scaling dynamics.

% \vspace{-7pt}
\subsubsection{Results and Analysis}

\textbf{Training Loss.} As shown in Fig.~\ref{fig:all_loss_curves}, our proposed methods achieve a consistent and significant reduction in training loss across all model scales and architectures. For the dense models, JTok consistently maintains a lower loss trajectory. Similarly, for the MoE models, both JTok and JTok-M show a clear advantage over the vanilla backbones, reducing the training loss for both the 1.5B and 3.2B MoE configurations. This shows that JTok and JTok-M are robust methods for improving optimization and data compression.

\textbf{Downstream Performance.} 
The benefits observed during training translate to significant and consistent improvements in downstream tasks. As shown in Table~\ref{tab:xl_moe_eval_all}, JTok boosts the Dense-XL average accuracy from 22.22 to 26.54 (+4.32), with sizable gains on key benchmarks such as MMLU (+4.55) and TriviaQA (+9.50). For MoE, gains are further amplified by JTok-M (Table~\ref{tab:xl_moe_eval_all}). On 1.5B-A250M, the average accuracy improves from 18.87 to 22.78 (+3.91); on 3.2B-A0.5B, it rises from 26.75 to 32.34 (+5.59), including strong improvements on ARC-C (+7.25) and GSM8K (+6.31).

\begin{figure*}[t]
   \centering
   \begin{subfigure}[b]{0.32\textwidth}
      \includegraphics[width=\textwidth]{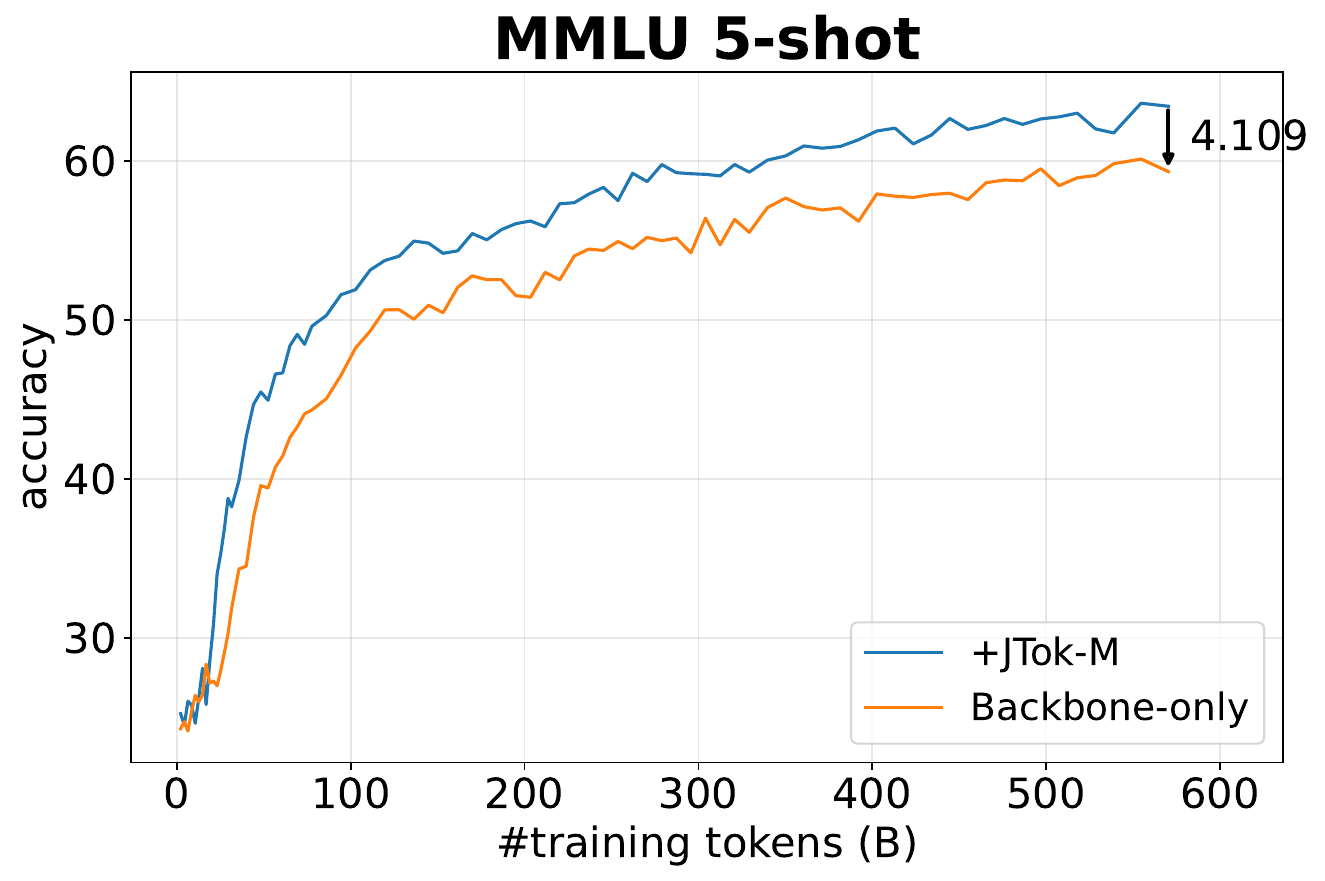}
      % \caption{MMLU accuracy}
      \label{fig:mmlu16b} 
   \end{subfigure}
   \hfill
   \begin{subfigure}[b]{0.32\textwidth}
      \includegraphics[width=\textwidth]{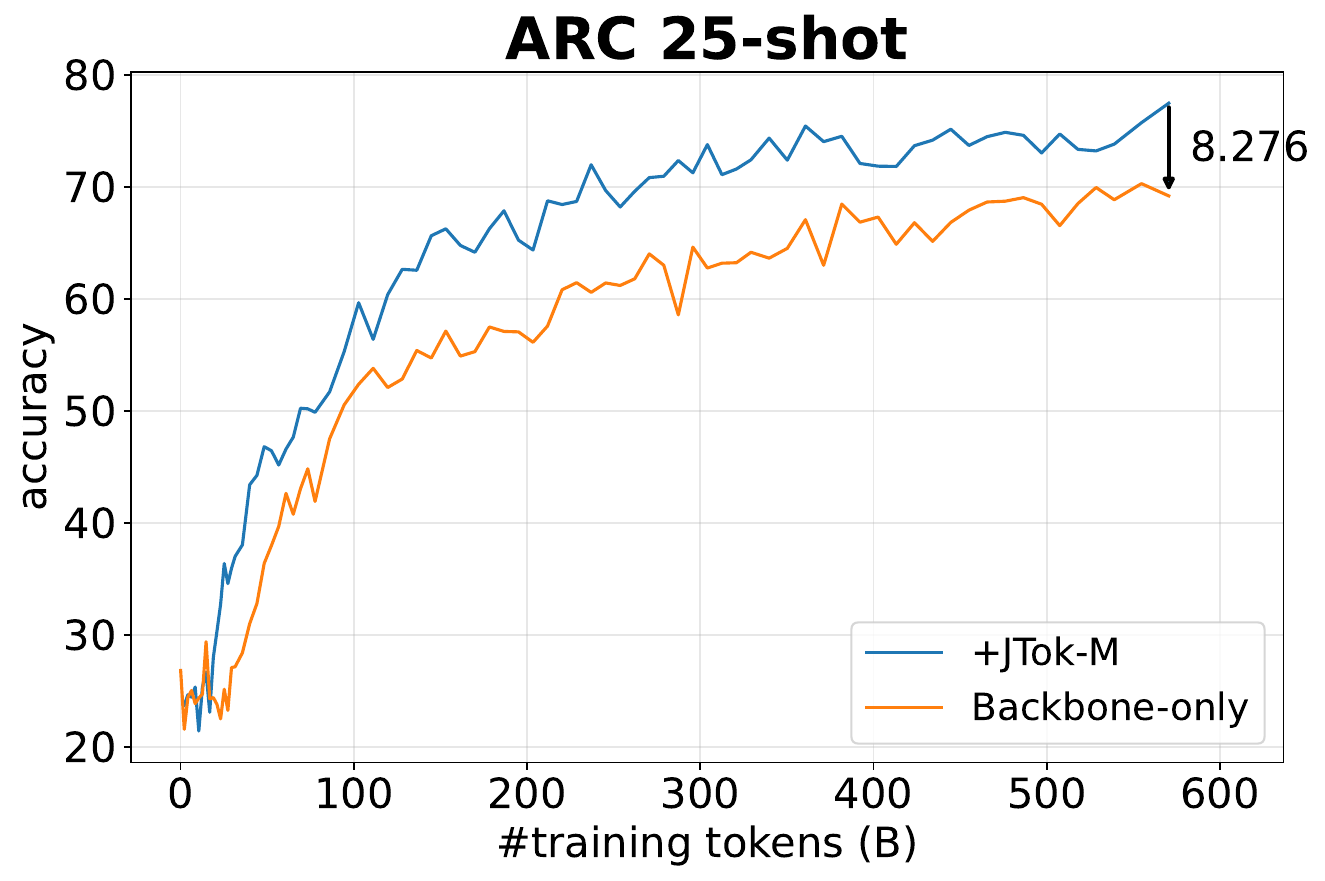}
      % \caption{ARC accuracy}
      \label{fig:arc16b}
   \end{subfigure}
   \hfill
   \begin{subfigure}[b]{0.32\textwidth}
      \includegraphics[width=\textwidth]{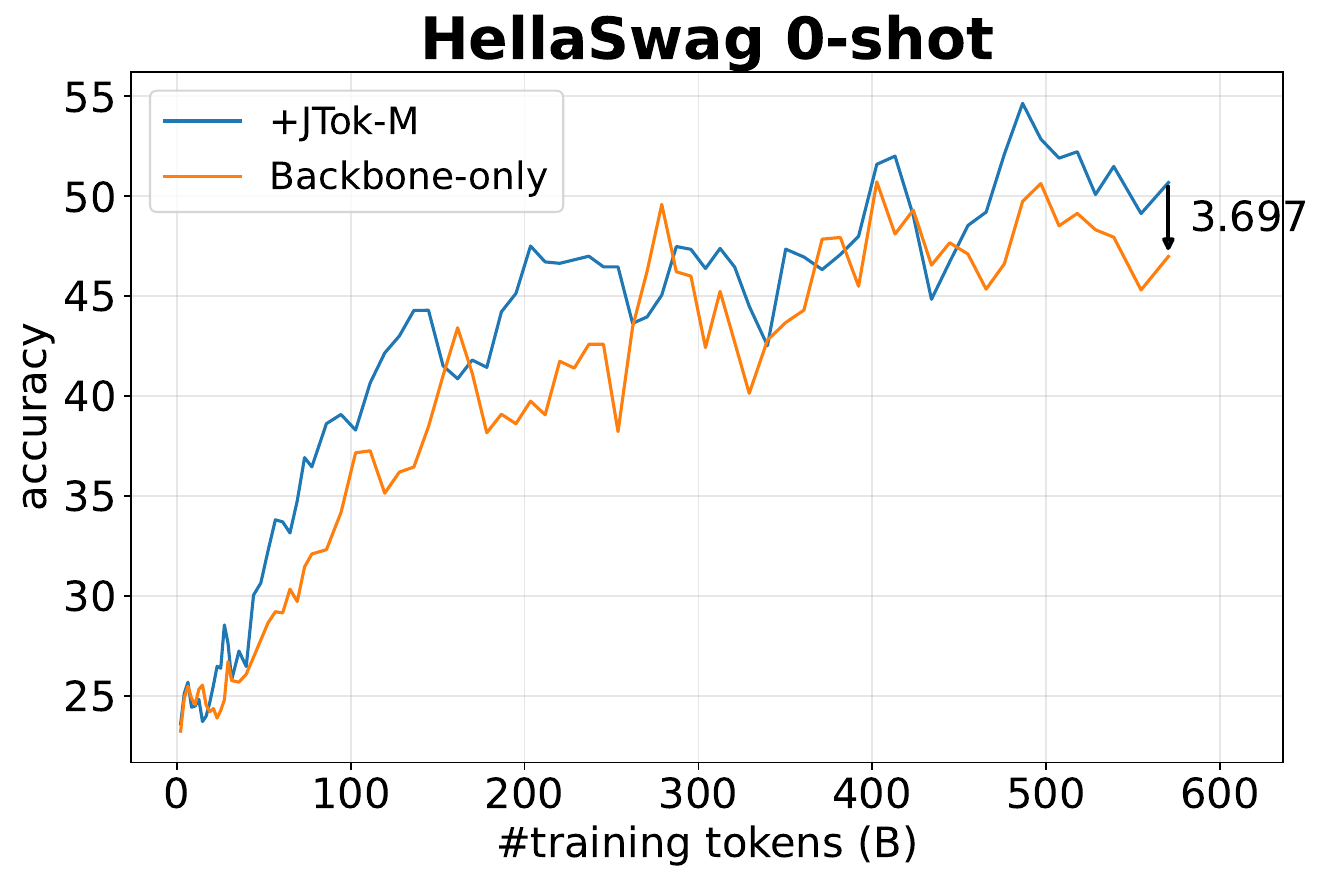}
      % \caption{Hellaswag accuracy}
      \label{fig:hellaswag16b}
   \end{subfigure}
   
   % \vspace{5pt} % Add some vertical space between rows
   
   \begin{subfigure}[b]{0.32\textwidth}
      \includegraphics[width=\textwidth]{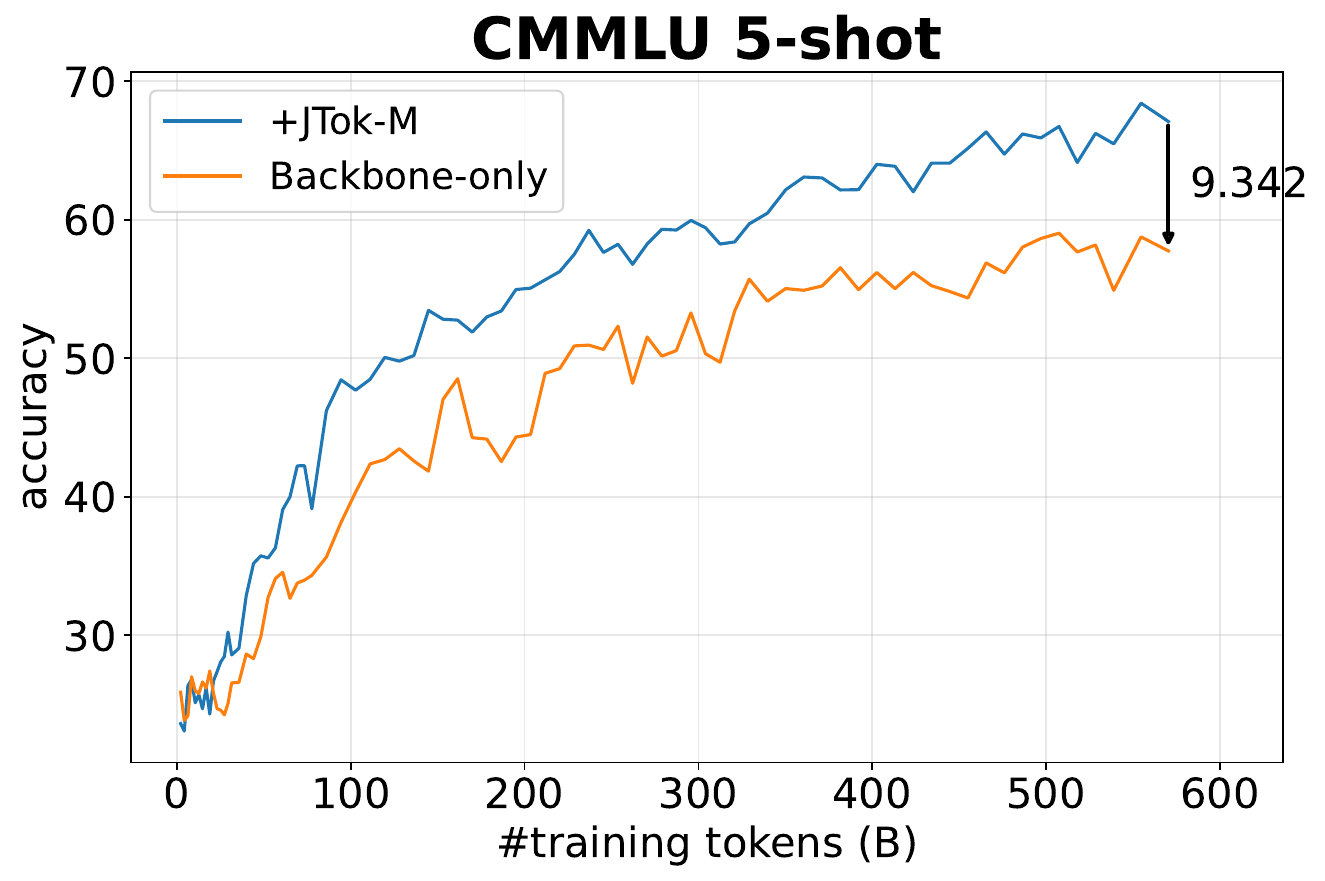}
      % \caption{CMMLU accuracy}
      \label{fig:cmmlu16b}
   \end{subfigure}
   \hfill
   \begin{subfigure}[b]{0.32\textwidth}
      \includegraphics[width=\textwidth]{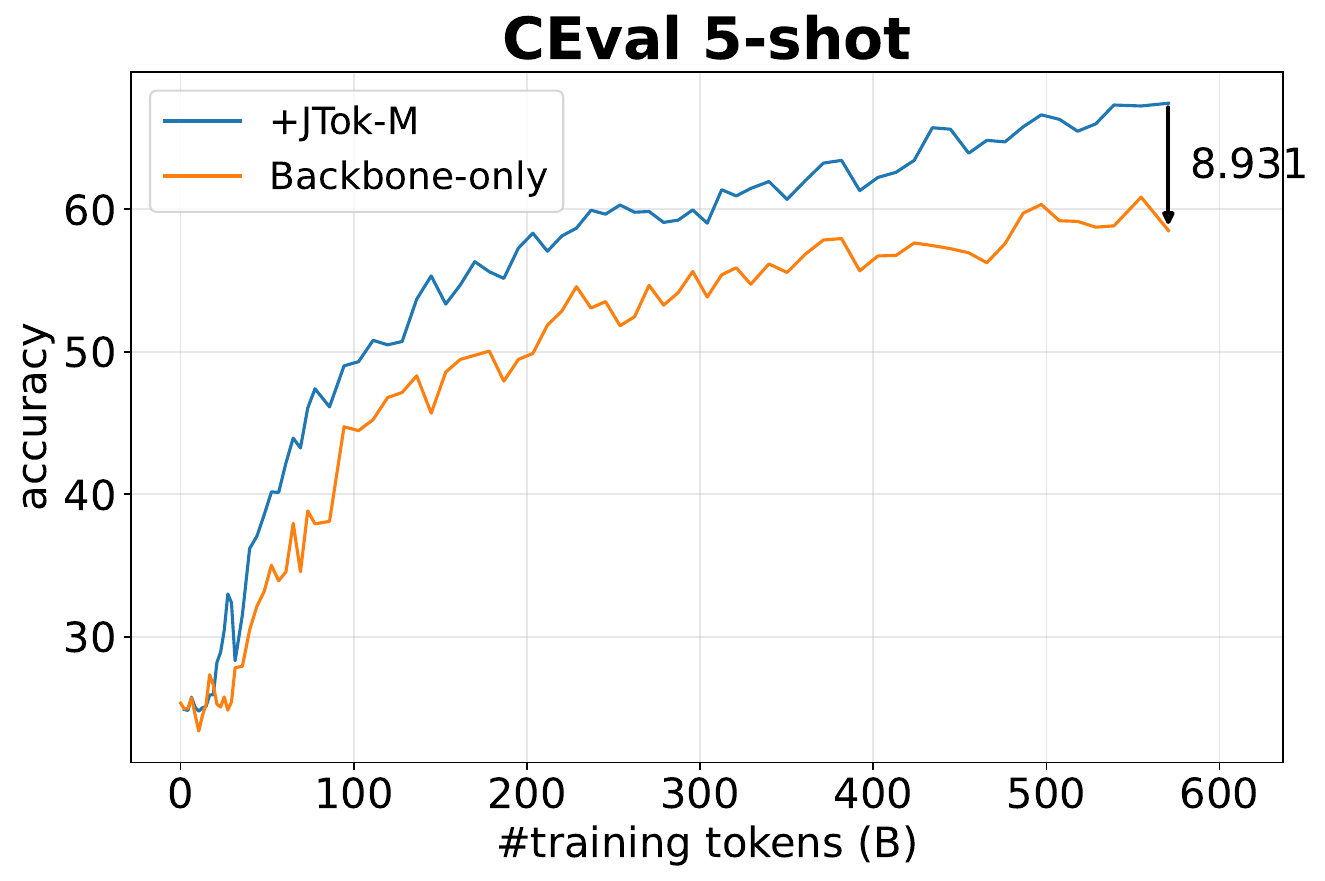} % Placeholder filename
      % \caption{CEval accuracy}
      \label{fig:ceval16b}
   \end{subfigure}
   \hfill
   \begin{subfigure}[b]{0.32\textwidth}
      \includegraphics[width=\textwidth]{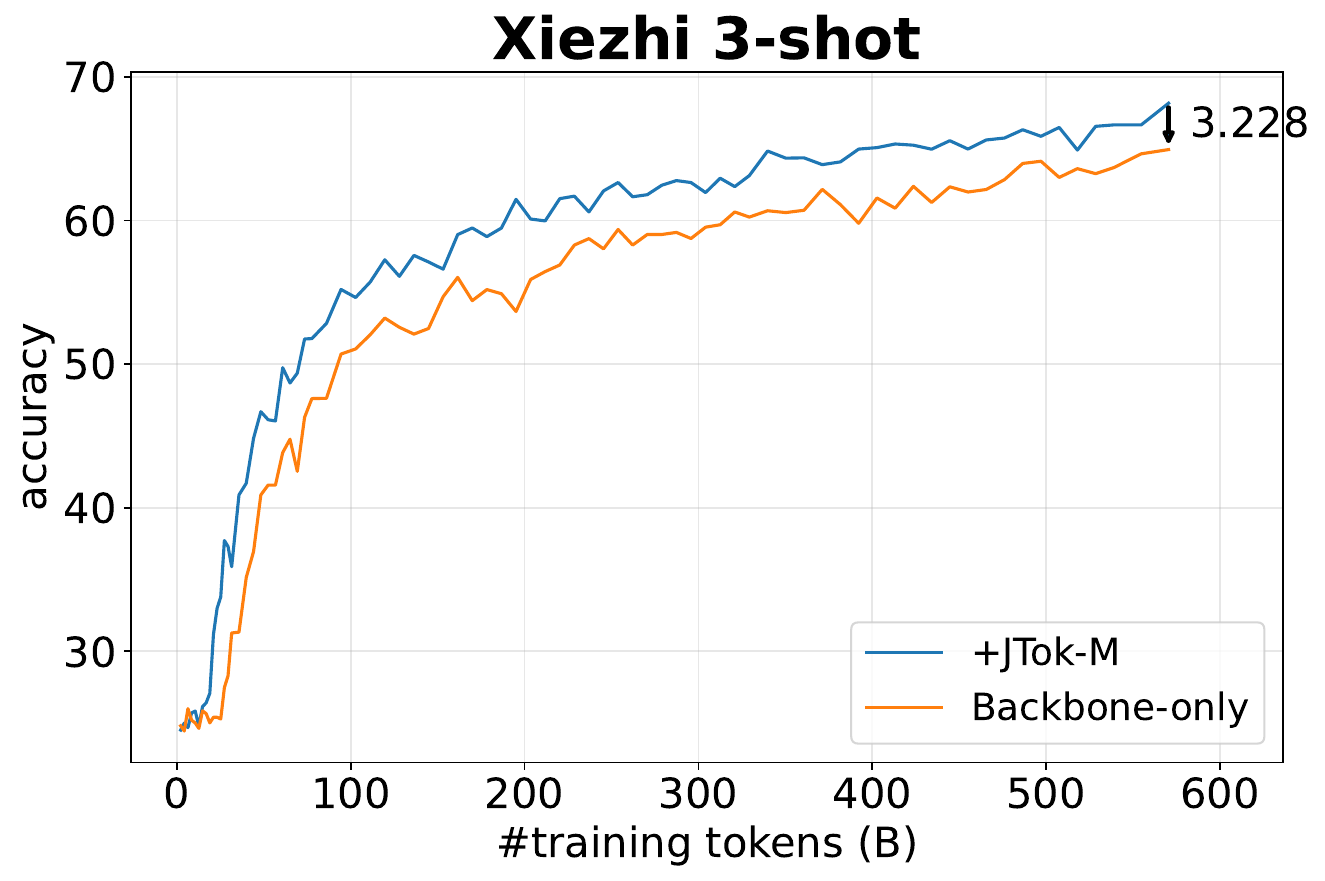} % Placeholder filename
      % \caption{CEval accuracy}
      \label{fig:xiezhi16b}
   \end{subfigure}
   
   \vspace{-5pt}
   \caption{\textbf{Downstream accuracy trajectories on the 17B-A2B MoE backbone.} We track the zero/few-shot performance on six representative benchmarks (MMLU, ARC-C, HellaSwag, CMMLU, C-Eval, and Xiezhi) throughout the 570B-token pretraining course. JTok-M (blue) consistently surpasses the backbone baseline (orange) from the early stages, yielding substantial gains in both knowledge-intensive and reasoning tasks without saturation.}
   \label{fig:eval16b}
   \vspace{-10pt}
\end{figure*}

Crucially, these gains scale effectively to the large-scale 17B-A2B MoE backbone. As illustrated in Fig.~\ref{fig:eval16b}, JTok-M delivers substantial boosts across diverse domains, particularly in knowledge-intensive and reasoning benchmarks—ranging from +4.11 on MMLU to an impressive +9.34 on CMMLU and +8.28 on ARC-C—validating the method's efficacy on foundation-scale models.

\textbf{Training Dynamics.} Beyond final metrics, the accuracy trajectories on the 17B backbone (Fig.~\ref{fig:eval16b}) reveal a distinct convergence advantage. JTok-M establishes a performance lead in the early training stage and widens this gap throughout the 570B-token course without saturation. Unlike techniques that only refine convergence at the tail end, token-indexed modulation provides a consistent capacity expansion, allowing the model to absorb complex semantics more efficiently from the onset. This confirms that token-indexed parameters function as an orthogonal scaling dimension that fundamentally elevates the model's learning capability and data efficiency.

\begin{table*}[t]
\centering
\caption{\textbf{Performance evaluation on comprehensive downstream tasks (Knowledge, Code, Reasoning, and Math).} The number in parentheses after each benchmark indicates the number of few-shot examples used in evaluation. B.O. denotes backbone-only; +JTok and +JTok-M denote the same backbone augmented with the corresponding plugin.
Overall Avg is the mean over all 14 tasks.
The best result of each task (within the same backbone block) is highlighted in bold.}
\label{tab:xl_moe_eval_all}
% \vspace{-4pt}

\setlength{\tabcolsep}{4pt}
\renewcommand{\arraystretch}{1.15}

\begin{tabular}{l cc ccc ccc}
\toprule
\textbf{Task} &
\multicolumn{2}{c}{\textbf{1.5B (Dense)}} &
\multicolumn{3}{c}{\textbf{1.5B-A250M (MoE)}} &
\multicolumn{3}{c}{\textbf{3.2B-A0.5B (MoE)}} \\
\cmidrule(lr){2-3}\cmidrule(lr){4-6}\cmidrule(lr){7-9}
& \textbf{B.O.} & \textbf{+JTok}
& \textbf{B.O.} & \textbf{+JTok} & \textbf{+JTok-M}
& \textbf{B.O.} & \textbf{+JTok} & \textbf{+JTok-M} \\
\midrule

\multicolumn{9}{l}{\textbf{Knowledge}} \\
MMLU(5)     & 32.74 & \textbf{37.29} & 27.91 & 30.98 & \textbf{34.07} & 36.39 & 40.05 & \textbf{43.47} \\
TriviaQA(5) & 25.72 & \textbf{35.22} & 22.81 & 27.31 & \textbf{36.53} & 39.57 & 41.13 & \textbf{46.29} \\
ARC-C(25)   & 35.74 & \textbf{41.58} & 29.21 & 30.24 & \textbf{34.72} & 39.76 & 43.87 & \textbf{47.01} \\
GPQA(5)     & 25.60 & \textbf{29.80} & 25.59 & 28.28 & \textbf{30.03} & 29.47 & 28.79 & \textbf{31.70} \\
\midrule

\multicolumn{9}{l}{\textbf{Code}} \\
MBPP(3)      &  9.53 & \textbf{12.20} &  7.00 & 10.60 & \textbf{12.50} & 19.20 & 21.80 & \textbf{24.51} \\
HumanEval(0) &  9.15 & \textbf{ 9.76} & \textbf{10.57} &  9.96 & 10.55 & 16.46 & 18.90 & \textbf{19.89} \\
LCB(5)       &  0.87 & \textbf{ 2.61} &  1.57 &  1.45 & \textbf{ 1.64} &  2.52 &  4.70 & \textbf{ 5.96} \\
\midrule

\multicolumn{9}{l}{\textbf{Reasoning}} \\
Hellaswag(10) & 51.19 & \textbf{55.07} & 45.70 & 48.01 & \textbf{52.06} & 57.05 & 58.52 & \textbf{60.37} \\
C3(3)        & 39.34 & \textbf{47.83} & 27.95 & 29.86 & \textbf{30.22} & 36.84 & 44.05 & \textbf{49.01} \\
BBH(3)       & 21.04 & \textbf{24.12} & 18.96 & 18.82 & \textbf{20.66} & 21.87 & 24.74 & \textbf{28.58} \\
SocialIQA(5) & 41.64 & \textbf{45.97} & 34.44 & 36.39 & \textbf{38.51} & 42.63 & 44.83 & \textbf{47.15} \\
\midrule

\multicolumn{9}{l}{\textbf{Math}} \\
MATH(4)  &  2.17 & \textbf{ 3.36} &  1.96 &  3.04 & \textbf{ 4.37} &  7.16 &  8.30 & \textbf{10.42} \\
GSM8K(4) &  5.76 & \textbf{ 9.07} &  3.87 & \textbf{ 4.17} &  4.15 & 13.65 & 16.38 & \textbf{19.96} \\
DROP(3)  & 10.55 & \textbf{17.71} &  6.60 &  7.99 & \textbf{ 8.90} & 11.86 & 13.27 & \textbf{18.44} \\
\midrule

\textbf{Overall Avg.} & 22.22 & \textbf{26.54} & 18.87 & 20.51 & \textbf{22.78} & 26.75 & 29.24 & \textbf{32.34} \\
\bottomrule
\end{tabular}

\vspace{-6pt}
\end{table*}

% \vspace{-7pt}
\subsection{Scaling Laws}
\label{sec:exp:scaling}
% \vspace{-7pt}

This subsection studies JTok-M scalability along two axes: (i) whether JTok-M's gains remain stable as the backbone scales, which is crucial for applying this feature in real-world LLMs; and (ii) what scaling behavior does it exhibit if scaling JTok-M itself.
Together, these results characterize token-indexed parameters as a predictable and complementary scaling dimension.

% \vspace{-7pt}
\subsubsection{JTok-M is scalable to large backbones} \label{sec:scale1}
% \vspace{-7pt}

\textbf{Large-scale Pretrain.}

We train both MoE and +JTok-M variant with three backbone sizes (2.3B-A0.5B, 3.9B-A0.6B, and A9.8B-A1.4B) under a fixed JTok-M configuration $(\eta,\rho)=(50,0.25)$ for 100B tokens.

\begin{figure}[t!]
\centering
\includegraphics[width=0.7\linewidth]{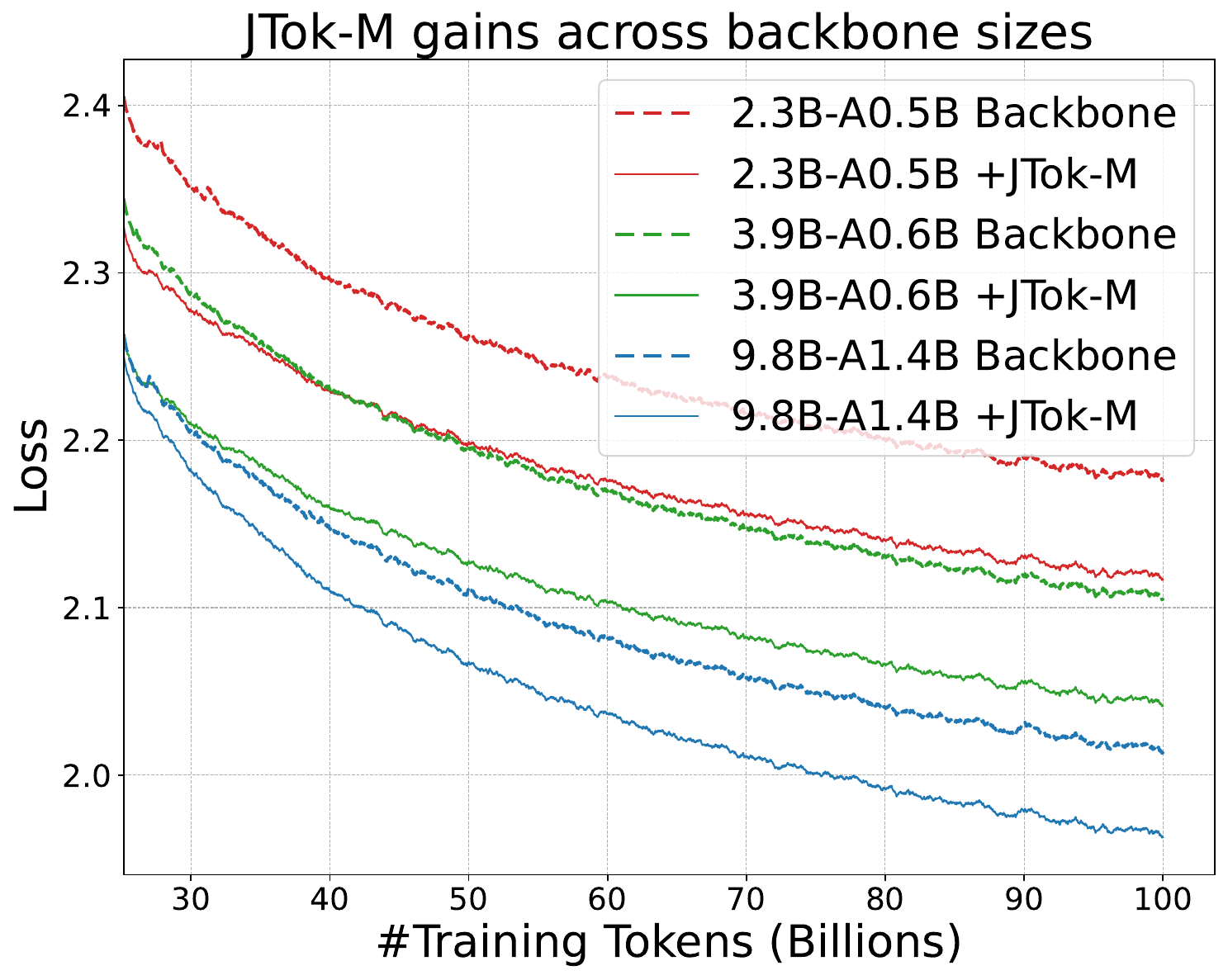}
\vspace{-7pt}
\caption{JTok-M (fixed $(\eta,\rho)$) improvements remain stable and consistent across different backbone model sizes.}
\label{fig:JTok-M_multiscale_loss_main}
% \vspace{-15pt}
\end{figure}

Fig.~\ref{fig:JTok-M_multiscale_loss_main} shows that JTok-M's benefits are remarkably consistent across model scales. JTok-M reduces the final loss by 0.059 on the 2.3B-A0.5B backbone, by 0.064 on the 3.9B-A0.6B backbone, and by 0.051 on the A9.8B-A1.4B backbone, corresponding to relative improvements of \textbf{2.7\%}, \textbf{3.0\%}, and \textbf{2.55\%}, respectively.

\textbf{Rigorous IsoFLOPs Validation}

In Sec.~\ref{method_scaling}, we predict that token-indexed parameters yield a \emph{scale-invariant} improvement under compute-optimal training: across compute budgets, the efficient frontier is shifted downward by a constant multiplicative factor (Eq.~\ref{eq:frontier_JTok-M}).
Here we demonstrate the validity of this hypothesis via isoFLOPs experiments.

\begin{figure}[tb!]
   \centering
  \includegraphics[width=.48\textwidth]{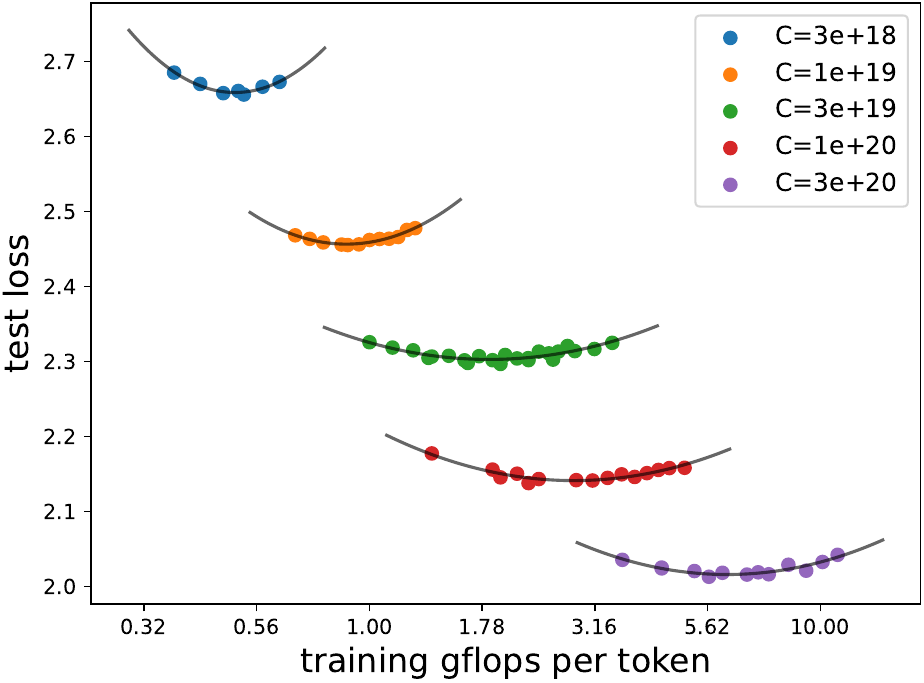}
 % \vspace{-8pt}
  \caption{\textbf{IsoFlOPs profile of vanilla MoE models.} Each curve shows test loss vs. per-token gFLOPs at fixed FLOPs, with optimal points defining the efficient frontier.}
  \label{fig:iso}
   
\end{figure}

Following the isoFLOPs protocal described in Sec.~\ref{iso-protocal}, we establish the efficient frontier for MoE (top-8 routing, one shared expert, and 145 total experts).
We choose five FLOPs budgets that are log-uniformly spaced, $C\in\{3e18,\,1e19,\,3e19,\,1e20,\,3e20\}$, and then fit a quadratic around the minimum of the resulting U-shaped curve to obtain the compute-optimal configuration $(N_c^*(C),D^*(C))$ and its best held-out loss $\mathcal{L}^*(C)$, as shown in Fig.~\ref{fig:iso}.

For each compute-optimal backbone configuration, we attach the JTok-M module while keeping the backbone architecture and training setup unchanged.
JTok-M adds token-indexed parameters but does not change the dominant FLOPs.
We use a fixed JTok-M configuration with parameter expansion ratio $\eta\!=\!N_n/N_c\!=\!50$ and activation sparsity $\rho\!=\!0.25$.
This yields another set of efficient-frontier points $\{(C,\mathcal{L}^*_{\mathrm{JTok-M}}(C))\}$ corresponding to JTok-M architecture.

\begin{figure*}[t]
   \centering
   \begin{subfigure}[t]{0.48\textwidth}
      \centering
      % \vspace{0pt}
      \includegraphics[width=\textwidth]{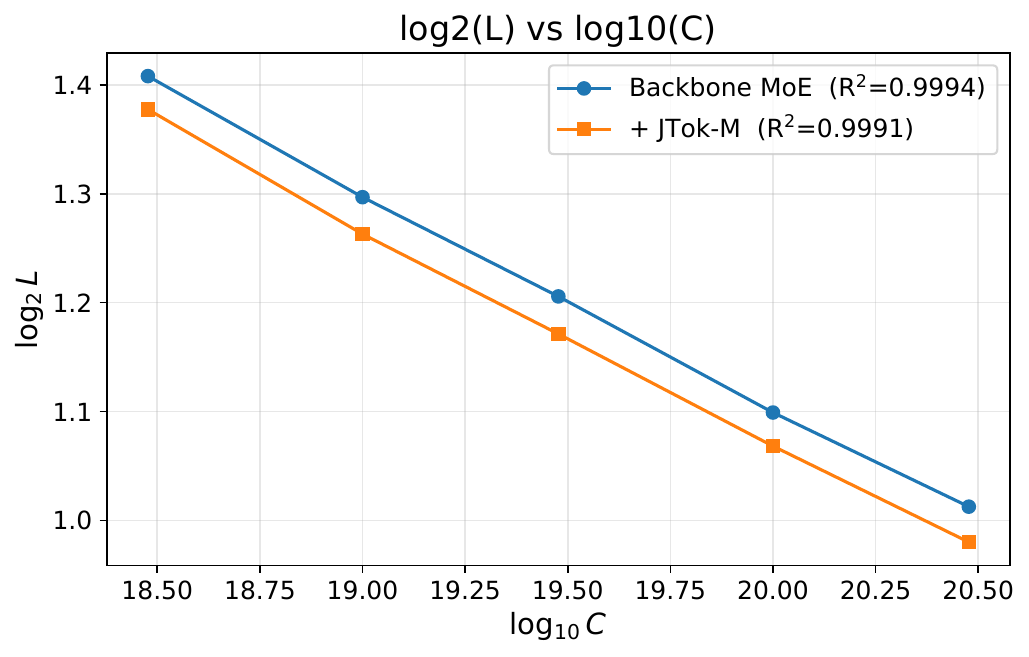}
      \caption{\textbf{Efficient frontiers of MoE and +JTok-M variant.} JTok-M achieves 2.2\% lower loss at each compute budget, equivalent to saving 35\% compute to reach each test loss.}
      \label{fig:ef} 
   \end{subfigure}
   \hfill
   \begin{subfigure}[t]{0.48\textwidth}
      \centering
      % \vspace{0pt}
      \includegraphics[width=\textwidth]{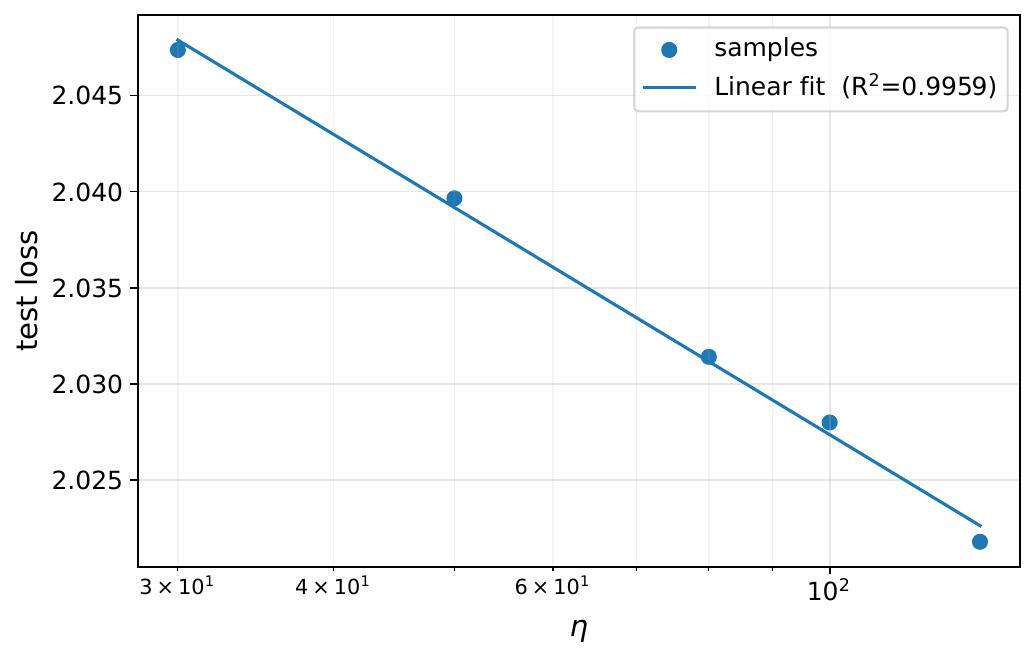}
      \caption{\textbf{Scaling behavior of JTok-M.} The validation loss follows an approximately log-linear trend with respect to the JTok-M parameter expansion ratio $\eta$.}
      \label{fig:JTok-M_scaling}
   \end{subfigure}
   \caption{JTok-M improves the efficient frontier and follows a log-linear scaling trend with $\eta$.}
   % \vspace{-7pt}
\end{figure*}

Across all budgets, JTok-M improves the compute-optimal loss, and the improvement is well-approximated by a constant multiplicative factor, consistent with Eq.~\ref{eq:frontier_JTok-M}.
To make the comparison explicit, we analyze the frontiers in log-log space.
Taking logarithms of Eq.~\ref{eq:frontier_JTok-M} gives
\begin{equation}
\label{eq:frontier_JTok-M_log}
\log \mathcal{L}^*_{\mathrm{JTok-M}}(C;\eta,\rho)
=
\log \mathcal{L}^*(C)-\frac{\alpha\beta}{\alpha+\beta}\,\log (1+\eta\gamma(\rho))
\end{equation}
which predicts that the vertical gap between the two frontiers in log space is $C$-invariant and depends solely on the JTok-M hyperparameters through $(\eta,\rho)$. Empirically, linear fits of $\log \mathcal{L}^*(C)$ and $\log \mathcal{L}^*_{\mathrm{JTok-M}}(C)$ versus $\log C$ yield nearly identical slopes, while JTok-M exhibits a clear downward intercept shift, as shown in Fig.~\ref{fig:ef}. 
This verifies that JTok-M improves the quality-compute Pareto frontier in a manner that is stable across scales. 
Quantitatively, under matched compute budgets, JTok-M achieves a consistent \textbf{2.2\% loss reduction}; equivalently, to reach any target test loss, JTok-M \textbf{saves 35\% compute} relative to MoE.
Detailed regression statistics and the mathematical derivation of the compute saving ratio are provided in Appendix~\ref{app:fit}.

\subsubsection{Scaling JTok-M itself}
In subsection~\ref{sec:scale1}, we validate that JTok-M can be reliably applied to larger backbone as its relative gains remain stable under large-scale pretraining, and the rigorous IsoFLOPs study verifies a scale-invariant downward shift of the compute-optimal frontier. We now examine the complementary question: \emph{can JTok-M itself be scaled as a standalone capacity axis, and does it exhibit a predictable scaling behavior?}

\textbf{Experimental Setup.}
We fix the backbone to a single MoE model (3.9B-A0.6B) and pretrain for 100B tokens. On top of this frozen backbone specification, we attach JTok-M modules of varying capacity. Concretely, we parameterize the JTok-M size using the parameter expansion ratio $\eta$; We sweep $\eta$ from 30 to 130 while fixing the sparsity ratio $\rho=0.25$.
All other JTok-M hyperparameters and all training hyperparameters are kept identical across runs, so that the observed differences are attributable solely to scaling $N_n$.

\textbf{Results}
Fig.~\ref{fig:JTok-M_scaling} shows that increasing JTok-M capacity leads to a clear and consistent reduction in held-out loss. Within the explored range, the improvement is monotonic with no sign of degradation, indicating that JTok-M provides an effective capacity knob beyond backbone scaling. Moreover, the trend is well-approximated by a smooth power law (linear in log-space): each doubling of JTok-M parameters reduces the loss by approximately $\mathbf{0.0118}$. This provides evidence that token-indexed parameters form an orthogonal scaling axis that introduces negligible backbone compute.

\begin{table}[tb!]
\centering

\caption{Training throughput (tokens/s) on the 3.2B-A0.5B MoE backbone with JTok-M ($\eta=50$). EmbP denotes embedding parallel, and Dedup denotes token deduplication.}
\label{tab:train_throughput}
\small
\setlength{\tabcolsep}{6pt}
\renewcommand{\arraystretch}{1.05}
\begin{tabular}{l c}
\toprule
\textbf{Model} & \textbf{Throughput (tok/s)} \\
\midrule
Baseline & 4,838K \\
JTok-M (naive) & 2,749K  \\
JTok-M (EmbP) & 4,024K \\
JTok-M (EmbP+Dedup) & 4,510K (-6.78\%)\\
\bottomrule
\end{tabular}

\vspace{6pt}

\caption{Inference throughput on 8$\times$H800 using SGLang (batch size=16, sequence length=4K) on the 3.2B-A0.5B MoE backbone. JTok/JTok-M use CPU offloading for token-indexed tables.}
\label{tab:infer_throughput}
\small
\setlength{\tabcolsep}{6pt}
\renewcommand{\arraystretch}{1.05}
\begin{tabular}{l c c}
\toprule
\textbf{Model} & \textbf{Prefill (tok/s)} & \textbf{Decode (tok/s)} \\
\midrule
Baseline & 363.7K & 4494 \\
JTok (CPU-offload) & 360.8K (-0.8\%) & 4290 (-4.5\%) \\
JTok-M (CPU-offload) & 355.2K (-2.3\%) & 4166 (-7.3\%)\\
\bottomrule
\end{tabular}

\end{table}
\subsection{System Efficiency} \label{sec:exp:efficiency}
% \vspace{-7pt}

We evaluate the end-to-end training and inference efficiency of JTok/JTok-M. With sufficient system optimizations, the throughput drop of JTok/JTok-M can be bounded within 6.78\% for training and 7.3\% for inference, demonstrating that JTok/JTok-M is practical for both large-scale training and deployment.

\textbf{Experimental Setup.}
(1) Training. We measure pretraining throughput (tokens/s) on 128$\times$H800, with batch size 1024 and sequence length 8192, on the 3.2B-A0.5B MoE backbone. For JTok-M, we set $\eta=50$ and $\rho=0.25$.
We compare (i) backbone-only, (ii) +JTok-M without optimization, (iii) +JTok-M with embedding parallelism, and (iv) +JTok-M with embedding parallelism plus token deduplication. \\
(2) Inference. We benchmark inference on 8$\times$H800 with batch size 16 and sequence length 4K using SGLang~\cite{sglang}. We implement CPU-offloaded JTok/JTok-M and report prefill and decode throughput on the 3.2B-A0.5B MoE backbone.

\textbf{Results.}
Table~\ref{tab:train_throughput} shows that a naive JTok-M implementation reduces training throughput mainly for two reasons: (i) the additional memory lookup and (ii) the token-indexed parameters increasing HBM usage, forcing a smaller micro-batch and lower MFU. Embedding parallelism alleviates the latter by sharding tables across GPUs to reduce per-device memory footprint, while token deduplication addresses the former by removing repeated gathers for frequent token access. With joint optimizations, the training throughput reduction is controlled within 6.78\%.

Table~\ref{tab:infer_throughput} shows that CPU-offloaded JTok/JTok-M incurs only small inference overhead. Prefill throughput drops by $0.8\%$ (JTok) and $2.3\%$ (JTok-M), while decode drops by $4.5\%$ and $7.3\%$, respectively. This is expected as decode is more compute-light than prefill, leaving less backbone computation to overlap and hide the host-to-device transfer latency. Overall, the throughput reduction remains low, supporting practical deployment.

% \vspace{-7pt}
\section{Conclusion and Outlook}
% \vspace{-7pt}

We have established token-indexed parameters as a novel and complementary scaling axis for large language models. Our proposed architectures, JTok and JTok-M, consistently enhance both dense and MoE models with negligible system overhead. Extensive evaluations show that our methods significantly boost performance across a wide range of downstream tasks.

Rigorous isoFLOPs analysis confirms that JTok-M fundamentally shifts the quality-compute Pareto frontier, achieving baseline-matching performance with 35\% compute savings. Furthermore, we empirically show that token-indexed parameter itself exhibits a predictable power-law scaling behavior, establishing this dimension as a robust and scalable trajectory for future LLM advancement.

\newpage

\bibliography{iclr2026_conference}

@article{vaswani2017attention,
  title={Attention is all you need},
  author={Vaswani, Ashish and Shazeer, Noam and Parmar, Niki and Uszkoreit, Jakob and Jones, Llion and Gomez, Aidan N and Kaiser, {\L}ukasz and Polosukhin, Illia},
  journal={Advances in neural information processing systems},
  volume={30},
  year={2017}
}

@article{kaplan2020scaling,
  title={Scaling laws for neural language models},
  author={Kaplan, Jared and McCandlish, Sam and Henighan, Tom and Brown, Tom B and Chess, Benjamin and Child, Rewon and Gray, Scott and Radford, Alec and Wu, Jeffrey and Amodei, Dario},
  journal={arXiv preprint arXiv:2001.08361},
  year={2020}
}

@article{hoffmann2022training,
  title={Training compute-optimal large language models (2022)},
  author={Hoffmann, Jordan and Borgeaud, Sebastian and Mensch, Arthur and Buchatskaya, Elena and Cai, Trevor and Rutherford, Eliza and de Las Casas, Diego and Hendricks, Lisa Anne and Welbl, Johannes and Clark, Aidan and others},
  journal={arXiv preprint arXiv:2203.15556},
  year={2022}
}

@article{fedus2022switch,
  title={Switch transformers: Scaling to trillion parameter models with simple and efficient sparsity},
  author={Fedus, William and Zoph, Barret and Shazeer, Noam},
  journal={Journal of Machine Learning Research},
  volume={23},
  number={120},
  pages={1--39},
  year={2022}
}

@article{deepseekmoe,
  title={Deepseekmoe: Towards ultimate expert specialization in mixture-of-experts language models},
  author={Dai, Damai and Deng, Chengqi and Zhao, Chenggang and Xu, RX and Gao, Huazuo and Chen, Deli and Li, Jiashi and Zeng, Wangding and Yu, Xingkai and Wu, Yu and others},
  journal={arXiv preprint arXiv:2401.06066},
  year={2024}
}

@article{kim2025pre,
  title={Pre-training under infinite compute},
  author={Kim, Konwoo and Kotha, Suhas and Liang, Percy and Hashimoto, Tatsunori},
  journal={arXiv preprint arXiv:2509.14786},
  year={2025}
}

@inproceedings{willwerunoutofdata,
  title={Position: Will we run out of data? Limits of LLM scaling based on human-generated data},
  author={Villalobos, Pablo and Ho, Anson and Sevilla, Jaime and Besiroglu, Tamay and Heim, Lennart and Hobbhahn, Marius},
  booktitle={Forty-first International Conference on Machine Learning},
  year={2024}
}

@article{tian2025towards,
  title={Towards greater leverage: Scaling laws for efficient mixture-of-experts language models},
  author={Tian, Changxin and Chen, Kunlong and Liu, Jia and Liu, Ziqi and Zhang, Zhiqiang and Zhou, Jun},
  journal={arXiv preprint arXiv:2507.17702},
  year={2025}
}

@article{huang2024toward,
  title={Toward efficient inference for mixture of experts},
  author={Huang, Haiyang and Ardalani, Newsha and Sun, Anna and Ke, Liu and Lee, Hsien-Hsin S and Bhosale, Shruti and Wu, Carole-Jean and Lee, Benjamin},
  journal={Advances in Neural Information Processing Systems},
  volume={37},
  pages={84033--84059},
  year={2024}
}

@article{deepseekv3,
  title={Deepseek-v3 technical report},
  author={Liu, Aixin and Feng, Bei and Xue, Bing and Wang, Bingxuan and Wu, Bochao and Lu, Chengda and Zhao, Chenggang and Deng, Chengqi and Zhang, Chenyu and Ruan, Chong and others},
  journal={arXiv preprint arXiv:2412.19437},
  year={2024}
}

@inproceedings{kim2024scaling,
  title={Scaling beyond the GPU memory limit for large mixture-of-experts model training},
  author={Kim, Yechan and Lim, Hwijoon and Han, Dongsu},
  booktitle={Forty-first International Conference on Machine Learning},
  year={2024}
}

@article{krajewski2024scaling,
  title={Scaling laws for fine-grained mixture of experts},
  author={Krajewski, Jakub and Ludziejewski, Jan and Adamczewski, Kamil and Pi{\'o}ro, Maciej and Krutul, Micha{\l} and Antoniak, Szymon and Ciebiera, Kamil and Kr{\'o}l, Krystian and Odrzyg{\'o}{\'z}d{\'z}, Tomasz and Sankowski, Piotr and others},
  journal={arXiv preprint arXiv:2402.07871},
  year={2024}
}

@article{krajewski2025scaling,
  title={Scaling Fine-Grained MoE Beyond 50B Parameters: Empirical Evaluation and Practical Insights},
  author={Krajewski, Jakub and Chochowski, Marcin and Korzekwa, Daniel},
  journal={arXiv preprint arXiv:2506.02890},
  year={2025}
}

@inproceedings{residual,
  title={Deep residual learning for image recognition},
  author={He, Kaiming and Zhang, Xiangyu and Ren, Shaoqing and Sun, Jian},
  booktitle={Proceedings of the IEEE conference on computer vision and pattern recognition},
  pages={770--778},
  year={2016}
}

@article{sardana2401beyond,
  title={Beyond chinchilla-optimal: Accounting for inference in language model scaling laws},
  author={Sardana, Nikhil and Portes, Jacob and Doubov, Sasha and Frankle, Jonathan},
  journal={arXiv preprint arXiv:2401.00448},
  year={2023}
}

@article{tao2024scaling,
  title={Scaling laws with vocabulary: Larger models deserve larger vocabularies},
  author={Tao, Chaofan and Liu, Qian and Dou, Longxu and Muennighoff, Niklas and Wan, Zhongwei and Luo, Ping and Lin, Min and Wong, Ngai},
  journal={Advances in Neural Information Processing Systems},
  volume={37},
  pages={114147--114179},
  year={2024}
}

@article{takase2024large,
  title={Large vocabulary size improves large language models},
  author={Takase, Sho and Ri, Ryokan and Kiyono, Shun and Kato, Takuya},
  journal={arXiv preprint arXiv:2406.16508},
  year={2024}
}

@article{huang2025over,
  title={Over-tokenized transformer: Vocabulary is generally worth scaling},
  author={Huang, Hongzhi and Zhu, Defa and Wu, Banggu and Zeng, Yutao and Wang, Ya and Min, Qiyang and Zhou, Xun},
  journal={arXiv preprint arXiv:2501.16975},
  year={2025}
}

@article{shazeer2017outrageously,
  title={Outrageously large neural networks: The sparsely-gated mixture-of-experts layer},
  author={Shazeer, Noam and Mirhoseini, Azalia and Maziarz, Krzysztof and Davis, Andy and Le, Quoc and Hinton, Geoffrey and Dean, Jeff},
  journal={arXiv preprint arXiv:1701.06538},
  year={2017}
}

@article{deepseekv2,
  title={Deepseek-v2: A strong, economical, and efficient mixture-of-experts language model},
  author={Liu, Aixin and Feng, Bei and Wang, Bin and Wang, Bingxuan and Liu, Bo and Zhao, Chenggang and Dengr, Chengqi and Ruan, Chong and Dai, Damai and Guo, Daya and others},
  journal={arXiv preprint arXiv:2405.04434},
  year={2024}
}

@article{jie2025mixture,
  title={Mixture of Lookup Experts},
  author={Jie, Shibo and Tang, Yehui and Han, Kai and Li, Yitong and Tang, Duyu and Deng, Zhi-Hong and Wang, Yunhe},
  journal={arXiv preprint arXiv:2503.15798},
  year={2025}
}

@article{yu2025scaling,
  title={Scaling embedding layers in language models},
  author={Yu, Da and Cohen, Edith and Ghazi, Badih and Huang, Yangsibo and Kamath, Pritish and Kumar, Ravi and Liu, Daogao and Zhang, Chiyuan},
  journal={arXiv preprint arXiv:2502.01637},
  year={2025}
}

@article{lample2019large,
  title={Large memory layers with product keys},
  author={Lample, Guillaume and Sablayrolles, Alexandre and Ranzato, Marc'Aurelio and Denoyer, Ludovic and J{\'e}gou, Herv{\'e}},
  journal={Advances in Neural Information Processing Systems},
  volume={32},
  year={2019}
}

@article{huang2024ultra,
  title={Ultra-sparse memory network},
  author={Huang, Zihao and Min, Qiyang and Huang, Hongzhi and Zhu, Defa and Zeng, Yutao and Guo, Ran and Zhou, Xun},
  journal={arXiv preprint arXiv:2411.12364},
  year={2024}
}

@article{berges2024memory,
  title={Memory layers at scale},
  author={Berges, Vincent-Pierre and O{\u{g}}uz, Barlas and Haziza, Daniel and Yih, Wen-tau and Zettlemoyer, Luke and Ghosh, Gargi},
  journal={arXiv preprint arXiv:2412.09764},
  year={2024}
}

@article{he2024mixture,
  title={Mixture of a million experts},
  author={He, Xu Owen},
  journal={arXiv preprint arXiv:2407.04153},
  year={2024}
}

@article{liu2025superbpe,
  title={Superbpe: Space travel for language models},
  author={Liu, Alisa and Hayase, Jonathan and Hofmann, Valentin and Oh, Sewoong and Smith, Noah A and Choi, Yejin},
  journal={arXiv preprint arXiv:2503.13423},
  year={2025}
}

@inproceedings{pagnoni2025byte,
  title={Byte latent transformer: Patches scale better than tokens},
  author={Pagnoni, Artidoro and Pasunuru, Ramakanth and Rodriguez, Pedro and Nguyen, John and Muller, Benjamin and Li, Margaret and Zhou, Chunting and Yu, Lili and Weston, Jason E and Zettlemoyer, Luke and others},
  booktitle={Proceedings of the 63rd Annual Meeting of the Association for Computational Linguistics (Volume 1: Long Papers)},
  pages={9238--9258},
  year={2025}
}

@article{rmsnorm,
  title={Root mean square layer normalization},
  author={Zhang, Biao and Sennrich, Rico},
  journal={Advances in neural information processing systems},
  volume={32},
  year={2019}
}

@article{prenorm,
  title={Learning deep transformer models for machine translation},
  author={Wang, Qiang and Li, Bei and Xiao, Tong and Zhu, Jingbo and Li, Changliang and Wong, Derek F and Chao, Lidia S},
  journal={arXiv preprint arXiv:1906.01787},
  year={2019}
}

@article{sigmoidgating,
  title={Sigmoid gating is more sample efficient than softmax gating in mixture of experts},
  author={Nguyen, Huy and Ho, Nhat and Rinaldo, Alessandro},
  journal={Advances in Neural Information Processing Systems},
  volume={37},
  pages={118357--118388},
  year={2024}
}

@article{radford2019language,
  title={Language models are unsupervised multitask learners},
  author={Radford, Alec and Wu, Jeffrey and Child, Rewon and Luan, David and Amodei, Dario and Sutskever, Ilya and others},
  journal={OpenAI blog},
  volume={1},
  number={8},
  pages={9},
  year={2019}
}

@article{qwen3next,
  title={Qwen3 technical report},
  author={Yang, An and Li, Anfeng and Yang, Baosong and Zhang, Beichen and Hui, Binyuan and Zheng, Bo and Yu, Bowen and Gao, Chang and Huang, Chengen and Lv, Chenxu and others},
  journal={arXiv preprint arXiv:2505.09388},
  year={2025}
}

@article{zhu2024hyper,
  title={Hyper-connections},
  author={Zhu, Defa and Huang, Hongzhi and Huang, Zihao and Zeng, Yutao and Mao, Yunyao and Wu, Banggu and Min, Qiyang and Zhou, Xun},
  journal={arXiv preprint arXiv:2409.19606},
  year={2024}
}

@misc{fineweb-edu,
    author       = { Lozhkov, Anton and Ben Allal, Loubna and von Werra, Leandro and Wolf, Thomas },  
    title        = { FineWeb-Edu: the Finest Collection of Educational Content }, 
    year         = 2024,  
    url          = { https://huggingface.co/datasets/HuggingFaceFW/fineweb-edu },  
    doi          = { 10.57967/hf/2497 },
    publisher    = { Hugging Face }
}

@misc{opencompass,
    title={OpenCompass: A Universal Evaluation Platform for Foundation Models},
    author={OpenCompass Contributors},
    howpublished = {\url{https://github.com/open-compass/opencompass}},
    year={2023}
}

@article{megatron-lm,
  title={Megatron-LM: Training Multi-Billion Parameter Language Models Using Model Parallelism},
  author={Shoeybi, Mohammad and Patwary, Mostofa and Puri, Raul and LeGresley, Patrick and Casper, Jared and Catanzaro, Bryan},
  journal={arXiv preprint arXiv:1909.08053},
  year={2019}
}

@article{mmlu,
  title={Measuring massive multitask language understanding},
  author={Hendrycks, Dan and Burns, Collin and Basart, Steven and Zou, Andy and Mazeika, Mantas and Song, Dawn and Steinhardt, Jacob},
  journal={arXiv preprint arXiv:2009.03300},
  year={2020}
}

@article{triviaqa,
  title={Triviaqa: A large scale distantly supervised challenge dataset for reading comprehension},
  author={Joshi, Mandar and Choi, Eunsol and Weld, Daniel S and Zettlemoyer, Luke},
  journal={arXiv preprint arXiv:1705.03551},
  year={2017}
}

@article{arcC,
  title={Think you have solved question answering? try arc, the ai2 reasoning challenge},
  author={Clark, Peter and Cowhey, Isaac and Etzioni, Oren and Khot, Tushar and Sabharwal, Ashish and Schoenick, Carissa and Tafjord, Oyvind},
  journal={arXiv preprint arXiv:1803.05457},
  year={2018}
}

@inproceedings{gpqa,
  title={Gpqa: A graduate-level google-proof q\&a benchmark},
  author={Rein, David and Hou, Betty Li and Stickland, Asa Cooper and Petty, Jackson and Pang, Richard Yuanzhe and Dirani, Julien and Michael, Julian and Bowman, Samuel R},
  booktitle={First Conference on Language Modeling},
  year={2024}
}

@article{c3,
  title={Investigating prior knowledge for challenging chinese machine reading comprehension},
  author={Sun, Kai and Yu, Dian and Yu, Dong and Cardie, Claire},
  journal={Transactions of the Association for Computational Linguistics},
  volume={8},
  pages={141--155},
  year={2020},
  publisher={MIT Press One Rogers Street, Cambridge, MA 02142-1209, USA journals-info~…}
}

@article{hellaswag,
  title={Hellaswag: Can a machine really finish your sentence?},
  author={Zellers, Rowan and Holtzman, Ari and Bisk, Yonatan and Farhadi, Ali and Choi, Yejin},
  journal={arXiv preprint arXiv:1905.07830},
  year={2019}
}

@article{bbh,
  title={Challenging big-bench tasks and whether chain-of-thought can solve them},
  author={Suzgun, Mirac and Scales, Nathan and Sch{\"a}rli, Nathanael and Gehrmann, Sebastian and Tay, Yi and Chung, Hyung Won and Chowdhery, Aakanksha and Le, Quoc V and Chi, Ed H and Zhou, Denny and others},
  journal={arXiv preprint arXiv:2210.09261},
  year={2022}
}

@article{socialiqa,
  title={Socialiqa: Commonsense reasoning about social interactions},
  author={Sap, Maarten and Rashkin, Hannah and Chen, Derek and LeBras, Ronan and Choi, Yejin},
  journal={arXiv preprint arXiv:1904.09728},
  year={2019}
}

@article{mbpp,
  title={Program synthesis with large language models},
  author={Austin, Jacob and Odena, Augustus and Nye, Maxwell and Bosma, Maarten and Michalewski, Henryk and Dohan, David and Jiang, Ellen and Cai, Carrie and Terry, Michael and Le, Quoc and others},
  journal={arXiv preprint arXiv:2108.07732},
  year={2021}
}

@article{humaneval,
  title={Evaluating large language models trained on code},
  author={Chen, Mark and Tworek, Jerry and Jun, Heewoo and Yuan, Qiming and Pinto, Henrique Ponde De Oliveira and Kaplan, Jared and Edwards, Harri and Burda, Yuri and Joseph, Nicholas and Brockman, Greg and others},
  journal={arXiv preprint arXiv:2107.03374},
  year={2021}
}

@article{livecodebench,
  title={Livecodebench: Holistic and contamination free evaluation of large language models for code},
  author={Jain, Naman and Han, King and Gu, Alex and Li, Wen-Ding and Yan, Fanjia and Zhang, Tianjun and Wang, Sida and Solar-Lezama, Armando and Sen, Koushik and Stoica, Ion},
  journal={arXiv preprint arXiv:2403.07974},
  year={2024}
}

@article{math,
  title={Measuring mathematical problem solving with the math dataset},
  author={Hendrycks, Dan and Burns, Collin and Kadavath, Saurav and Arora, Akul and Basart, Steven and Tang, Eric and Song, Dawn and Steinhardt, Jacob},
  journal={arXiv preprint arXiv:2103.03874},
  year={2021}
}

@article{gsm8k,
  title={Training verifiers to solve math word problems},
  author={Cobbe, Karl and Kosaraju, Vineet and Bavarian, Mohammad and Chen, Mark and Jun, Heewoo and Kaiser, Lukasz and Plappert, Matthias and Tworek, Jerry and Hilton, Jacob and Nakano, Reiichiro and others},
  journal={arXiv preprint arXiv:2110.14168},
  year={2021}
}

@article{dua2019drop,
  title={DROP: A reading comprehension benchmark requiring discrete reasoning over paragraphs},
  author={Dua, Dheeru and Wang, Yizhong and Dasigi, Pradeep and Stanovsky, Gabriel and Singh, Sameer and Gardner, Matt},
  journal={arXiv preprint arXiv:1903.00161},
  year={2019}
}

@inproceedings{xiezhi,
  title={Xiezhi: An ever-updating benchmark for holistic domain knowledge evaluation},
  author={Gu, Zhouhong and Zhu, Xiaoxuan and Ye, Haoning and Zhang, Lin and Wang, Jianchen and Zhu, Yixin and Jiang, Sihang and Xiong, Zhuozhi and Li, Zihan and Wu, Weijie and others},
  booktitle={Proceedings of the AAAI conference on artificial intelligence},
  volume={38},
  number={16},
  pages={18099--18107},
  year={2024}
}

@article{ceval,
  title={C-eval: A multi-level multi-discipline chinese evaluation suite for foundation models},
  author={Huang, Yuzhen and Bai, Yuzhuo and Zhu, Zhihao and Zhang, Junlei and Zhang, Jinghan and Su, Tangjun and Liu, Junteng and Lv, Chuancheng and Zhang, Yikai and Fu, Yao and others},
  journal={Advances in Neural Information Processing Systems},
  volume={36},
  pages={62991--63010},
  year={2023}
}

@inproceedings{cmmlu,
  title={Cmmlu: Measuring massive multitask language understanding in chinese},
  author={Li, Haonan and Zhang, Yixuan and Koto, Fajri and Yang, Yifei and Zhao, Hai and Gong, Yeyun and Duan, Nan and Baldwin, Timothy},
  booktitle={Findings of the Association for Computational Linguistics: ACL 2024},
  pages={11260--11285},
  year={2024}
}

@article{qwen3,
  title={Qwen3 technical report},
  author={Yang, An and Li, Anfeng and Yang, Baosong and Zhang, Beichen and Hui, Binyuan and Zheng, Bo and Yu, Bowen and Gao, Chang and Huang, Chengen and Lv, Chenxu and others},
  journal={arXiv preprint arXiv:2505.09388},
  year={2025}
}

@article{adam,
  title={Adam: A method for stochastic optimization},
  author={Kingma, Diederik P},
  journal={arXiv preprint arXiv:1412.6980},
  year={2014}
}

@article{roofline,
  title={Roofline: an insightful visual performance model for multicore architectures},
  author={Williams, Samuel and Waterman, Andrew and Patterson, David},
  journal={Communications of the ACM},
  volume={52},
  number={4},
  pages={65--76},
  year={2009},
  publisher={ACM New York, NY, USA}
}

@book{zipf,
  title={Human behavior and the principle of least effort: An introduction to human ecology},
  author={Zipf, George Kingsley},
  year={2016},
  publisher={Ravenio books}
}

@article{pearce2024reconciling,
  title={Reconciling Kaplan and Chinchilla Scaling Laws},
  author={Pearce, Tim and Song, Jinyeop},
  journal={Transactions on Machine Learning Research},
  year={2024}
}

@article{sglang,
  title={Sglang: Efficient execution of structured language model programs},
  author={Zheng, Lianmin and Yin, Liangsheng and Xie, Zhiqiang and Sun, Chuyue Livia and Huang, Jeff and Yu, Cody Hao and Cao, Shiyi and Kozyrakis, Christos and Stoica, Ion and Gonzalez, Joseph E and others},
  journal={Advances in neural information processing systems},
  volume={37},
  pages={62557--62583},
  year={2024}
}

@article{ngpt,
  title={ngpt: Normalized transformer with representation learning on the hypersphere},
  author={Loshchilov, Ilya and Hsieh, Cheng-Ping and Sun, Simeng and Ginsburg, Boris},
  journal={arXiv preprint arXiv:2410.01131},
  year={2024}
}

@article{wsd,
  title={Understanding warmup-stable-decay learning rates: A river valley loss landscape perspective},
  author={Wen, Kaiyue and Li, Zhiyuan and Wang, Jason and Hall, David and Liang, Percy and Ma, Tengyu},
  journal={arXiv preprint arXiv:2410.05192},
  year={2024}
}
\bibliographystyle{iclr2026_conference}

\newpage
\appendix

\section{Derivations for JTok-M scaling laws}
\label{app:derivation}

Here we provide detailed derivations for Eq.~\ref{eq:frontier_JTok-M} and Eq.~\ref{eq:compute_saving}.

\paragraph{Step 1: JTok-M as an effective rescaling of the model-size term.}
We start from the Kaplan form (Eq.~\ref{eq:kaplan}).
For JTok-M, we introduce the effective parameter count
\begin{equation}
N_{\mathrm{eff}}
= N_c+\gamma(\rho)N_n
= N_c\bigl(1+\eta\gamma(\rho)\bigr),
\qquad \eta\triangleq N_n/N_c.
\end{equation}
Plugging $N_{\mathrm{eff}}$ into Eq.~\ref{eq:kaplan} yields
\begin{align}
\mathcal{L}_{\mathrm{JTok-M}}(N_c,D;\eta,\rho)
&=
\Biggl[
\Bigl(\frac{A}{N_{\mathrm{eff}}}\Bigr)^{\frac{\alpha}{\beta}}
+\frac{B}{D}
\Biggr]^{\beta} \nonumber\\
&=
\Biggl[
\Bigl(\frac{A/(1+\eta\gamma(\rho))}{N_c}\Bigr)^{\frac{\alpha}{\beta}}
+\frac{B}{D}
\Biggr]^{\beta}.
\end{align}
Define
\begin{equation}
A_{\mathrm{JTok-M}}\triangleq \frac{A}{1+\eta\gamma(\rho)}.
\end{equation}
Then JTok-M has the same functional form as the backbone loss, except that the constant
in the model-size term is rescaled from $A$ to $A_{\mathrm{JTok-M}}$.

\paragraph{Step 2: Compute-optimal frontier and its dependence on $A$.}
Under the standard training-FLOPs approximation $C\approx 6N_cD$ (Sec.~\ref{iso-protocal}),
we can write $D=C/(6N_c)$ and view the loss as a function of $N_c$ at fixed $C$:
\begin{equation}
\mathcal{L}(N_c;C)
=
\Biggl[
\Bigl(\frac{A}{N_c}\Bigr)^{\frac{\alpha}{\beta}}
+\frac{6BN_c}{C}
\Biggr]^{\beta}.
\end{equation}
Minimizing the bracketed term over $N_c$ gives the compute-optimal frontier $\mathcal{L}^* (C)$.
A standard calculation for this two-term trade-off (capacity term decays with $N_c$ while data term
grows with $N_c$ under fixed $C$) yields the well-known dependence
\begin{equation}
\mathcal{L}^*(C)\ \propto\ A^{\frac{\alpha\beta}{\alpha+\beta}}\,C^{-\frac{\alpha\beta}{\alpha+\beta}},
\label{eq:frontier_base_scaling}
\end{equation}
i.e., at compute-optimality the frontier is a power law in $C$, and its \emph{intercept} scales as
$A^{\frac{\alpha\beta}{\alpha+\beta}}$.

Since JTok-M only changes $A\mapsto A_{\mathrm{JTok-M}}$ while keeping the same $C$-dependence,
we immediately obtain the multiplicative shift:
\begin{align}
\mathcal{L}_{\mathrm{JTok-M}}^*(C;\eta,\rho)
&=
\Bigl(\frac{A_{\mathrm{JTok-M}}}{A}\Bigr)^{\frac{\alpha\beta}{\alpha+\beta}}
\mathcal{L}^*(C) \nonumber\\
&=
\bigl(1+\eta\gamma(\rho)\bigr)^{-\frac{\alpha\beta}{\alpha+\beta}}
\cdot \mathcal{L}^*(C),
\end{align}
which is Eq.~(16).

\paragraph{Step 3: Isoperformance compute saving.}
Eq.~(16) implies that, in the compute-optimal regime, JTok-M improves the frontier by a \emph{constant}
multiplicative factor independent of $C$.
Let $k\triangleq \frac{\alpha\beta}{\alpha+\beta}$ and write $\mathcal{L}^*(C)=\kappa C^{-k}$ for some constant $\kappa$.
Then Eq.~(16) becomes
\begin{equation}
\mathcal{L}_{\mathrm{JTok-M}}^*(C)= (1+\eta\gamma(\rho))^{-k}\kappa C^{-k}.
\end{equation}
For any target loss $\mathcal{L}^\star$, solving $\mathcal{L}^\star=\kappa(C_{\mathrm{base}}^*)^{-k}$
and $\mathcal{L}^\star=(1+\eta\gamma(\rho))^{-k}\kappa(C_{\mathrm{JTok-M}}^*)^{-k}$ gives
\begin{equation}
C_{\mathrm{JTok-M}}^*(\mathcal{L}^\star)
=
\frac{1}{1+\eta\gamma(\rho)}\,C_{\mathrm{base}}^*(\mathcal{L}^\star),
\end{equation}
which is Eq.~(17). Notably, the compute saving ratio depends only on JTok-M hyperparameters
through $(\eta,\rho)$ and $\gamma(\rho)$, and is independent of the backbone scale.

\section{Load-Balancing Auxiliary Loss for JTok-M}
\label{app:loadbalance}

Similar to MoE architectures that require balanced expert utilization for optimal throughput and learning, JTok-M benefits from uniform routing across its $n_e$ embedding experts $\{E_i\}_{i=1}^{n_e}$. To encourage all embeddings are adequately trained and contribute to model performance, we incorporate an auxiliary load-balancing loss that encourages balanced routing distributions.

\textbf{Formulation.} Consider a training batch with $T$ tokens, where JTok-M maintains $n_e$ embedding experts and each token routes to top-$K$ embeddings. Let $G_t \subseteq \{1,\dots,n_e\}$ denote the top-$K$ routing set for token $t$, and $p_t^{(i)} \in [0,1]$ represent the routing probability of token $t$ to embedding $i$.

We define the following quantities:
\begin{itemize}[leftmargin=*, itemsep=0pt, topsep=0pt]
    \item \textbf{Aggregate routing probability:} $P_i = \sum_{t=1}^{T} p_t^{(i)}$ — the unnormalized sum of routing probabilities for embedding $i$
    \item \textbf{Actual token count:} $n_i = \sum_{t=1}^{T} \mathbf{1}\{i \in G_t\}$ — number of tokens actually routed to embedding $i$
    \item \textbf{Normalized routing probability:} $p_i = \frac{P_i}{T}$ — the average routing probability for embedding $i$
    \item \textbf{Normalized load fraction:} $f_i = \frac{n_i}{TK}$ — the fraction of total route capacity used by embedding $i$
\end{itemize}

The load-balancing auxiliary loss is formulated as:
\begin{equation}
    \mathcal{L}_{\mathrm{aux}} = \lambda \cdot n_e \sum_{i=1}^{n_e} p_i f_i = \lambda \cdot \frac{n_e}{T^2 K} \sum_{i=1}^{n_e} P_i n_i
    \label{eq:aux_loss}
\end{equation}
where $\lambda$ is a hyperparameter controlling the strength of the load-balancing constraint.

\textbf{Intuition.} The term $p_i$ captures the expected routing distribution to embedding $i$, while $f_i$ measures its actual utilization. The loss in Equation~\ref{eq:aux_loss} penalizes the co-occurrence of high routing probability and high actual load, thereby discouraging the concentration of routing on a subset of embeddings. This encourages a more uniform distribution where ideally $p_i \approx f_i \approx 1/n_e$ for all $i$, ensuring that all embeddings receive sufficient training signal and contribute effectively to model capacity.

\textbf{Implementation.} In practice, we compute this auxiliary loss per layer and average across all JTok-M layers. The hyperparameter $\lambda$ is typically set to $10^{-4}$, balancing load distribution without overwhelming the primary language modeling objective. This auxiliary loss is added to the main cross-entropy loss during training and is automatically handled by the backward pass without requiring special gradient computation.

\section{Detailed Compute-Optimal Scaling Laws} \label{app:fit}

In this section, we provide the detailed numerical results and linear fitting parameters for the IsoFLOPs analysis discussed in Sec.~\ref{sec:scale1}. We compare the compute-optimal performance of the vanilla MoE backbone against the JTok-M augmented variant ($\eta=50, \rho=0.25$) across five orders of magnitude of compute budgets.

\subsection{Compute-Optimal Data Points}

Table~\ref{tab:iso_flops_data} lists the compute-optimal test losses ($\mathcal{L}^*$) for both the vanilla MoE backbone and the JTok-M model at five distinct FLOPs budgets ($C$). These values correspond to the minima of the IsoFLOPs curves. The relative improvement is calculated as $1 - (\mathcal{L}^*_{\text{JTok-M}} / \mathcal{L}^*_{\text{Base}})$.

\begin{table}[h]
    \centering
    \caption{Compute-optimal test loss comparison between Vanilla MoE and JTok-M across different compute budgets. The data corresponds to the extracted efficient frontiers.}
    \label{tab:iso_flops_data}
    \setlength{\tabcolsep}{8pt}
    \begin{tabular}{ccccc}
        \toprule
        \textbf{Budget ($C$)} & \textbf{Vanilla MoE} & \textbf{JTok-M}  \\
        (FLOPs) & ($\mathcal{L}^*_{\text{Base}}$) & ($\mathcal{L}^*_{\text{JTok-M}}$) \\
        \midrule
        $3\times 10^{18}$ & 2.6537 & 2.5981  \\
        $1\times 10^{19}$ & 2.4569 & 2.3999\\
        $3\times 10^{19}$ & 2.3065 & 2.2521 \\
        $1\times 10^{20}$ & 2.1422 & 2.0969 \\
        $3\times 10^{20}$ & 2.0176 & 1.9726  \\
        \bottomrule
    \end{tabular}
\end{table}

\subsection{Linear Fitting of Efficient Frontiers}

To characterize the scaling behavior, we perform a linear regression in the log-log space. Following the standard practice, we use $\log_{10}$ for the compute budget $C$ and $\log_2$ for the loss $L$ (to align with previous work numerical scales). The relationship is modeled as:
\begin{equation}
    \log_2(\mathcal{L}) = \alpha \cdot \log_{10}(C) + \beta
\end{equation}

Based on the data points in Table~\ref{tab:iso_flops_data}, we obtain the following linear fits:

\paragraph{Vanilla MoE (Backbone)}
\begin{equation}
    \log_2(\mathcal{L}) = -0.2016 \cdot \log_{10}(C) + 5.1334, \quad R^2 = 0.9994
\end{equation}

\paragraph{JTok-M}
\begin{equation}
    \log_2(\mathcal{L}) = -0.2009 \cdot \log_{10}(C) + 5.0881, \quad R^2 = 0.9991
\end{equation}

\subsection{Derivation of Compute Saving ratio}
\label{app:compute_saving_derivation}

In this subsection, we provide the mathematical derivation for the $35\%$ compute saving claim based on the fitted efficient frontiers.

\paragraph{Formulation}
Let the scaling laws for the baseline and JTok-M models be expressed in the log-log space as:
\begin{align}
    \log_2(\mathcal{L}_{\text{base}}) &= \alpha \cdot \log_{10}(C) + \beta_{\text{base}} \\
    \log_2(\mathcal{L}_{\text{JTok-M}}) &= \alpha \cdot \log_{10}(C) + \beta_{\text{JTok-M}}
\end{align}
where $\alpha$ represents the scaling slope (empirically $\approx -0.2016$) and $\beta$ represents the intercept. We assume the slopes are identical based on the empirical fits shown in Fig.~6a.

\paragraph{Compute Ratio for Iso-Loss}
To estimate the compute saving, we determine the ratio of compute budgets ($C_{\text{JTok-M}}$ vs. $C_{\text{base}}$) required to achieve the same target validation loss $\mathcal{L}^*$. Setting $\mathcal{L}_{\text{base}}(C_{\text{base}}) = \mathcal{L}_{\text{JTok-M}}(C_{\text{JTok-M}}) = \mathcal{L}^*$:
\begin{equation}
    \alpha \cdot \log_{10}(C_{\text{base}}) + \beta_{\text{base}} = \alpha \cdot \log_{10}(C_{\text{JTok-M}}) + \beta_{\text{JTok-M}}
\end{equation}

Rearranging the terms to solve for the compute ratio:
\begin{align}
    \alpha (\log_{10}(C_{\text{JTok-M}}) - \log_{10}(C_{\text{base}})) &= \beta_{\text{base}} - \beta_{\text{JTok-M}} \\
    \log_{10}\left(\frac{C_{\text{JTok-M}}}{C_{\text{base}}}\right) &= \frac{\beta_{\text{base}} - \beta_{\text{JTok-M}}}{\alpha} = \frac{\Delta \beta}{\alpha}
\end{align}
Thus, the compute ratio is given by:
\begin{equation}
    \text{Ratio} = \frac{C_{\text{JTok-M}}}{C_{\text{base}}} = 10^{\frac{\Delta \beta}{\alpha}}
\end{equation}

\paragraph{Calculation}
Using the fitted parameters from our experiments:
\begin{itemize}
    \item Slope $\alpha \approx -0.2016$
    \item Intercept difference $\Delta \beta = \beta_{\text{base}} - \beta_{\text{JTok-M}} \approx 0.038$ 
\end{itemize}

Substituting these values:
\begin{equation}
    \text{Ratio} = 10^{\frac{0.038}{-0.2016}} \approx 10^{-0.1885} \approx 0.648
\end{equation}

Finally, the compute saving is:
\begin{equation}
    \text{Saving} = 1 - \text{Ratio} = 1 - 0.648 \approx 35.2\%
\end{equation}
This confirms that JTok-M reduces the compute budget required to reach a target loss by approximately $35\%$.

\section{Ablation Studies}

\subsection{Scaling factor $\frac{1}{\sqrt{2N_l}}$ in JTok-M} \label{app:ablation_downscaling}

To stabilize the variance of hidden states when injecting JTok-M residuals, we scale the modulation update in each layer with a factor $\frac{1}{\sqrt{2N_l}}$ (Eq.~\ref{eq_down_scale}).

We perform an ablation on the 3.2B–A0.5B MoE backbone comparing (i) the baseline(backbone MoE), (ii) JTok-M \emph{with} the scaling factor (our default), and (iii) JTok-M \emph{without} the scaling factor. All models are trained under identical data and optimization settings.

\begin{figure}[t!]
\centering
\begin{subfigure}[b]{0.48\textwidth}
    \centering
    \includegraphics[width=\textwidth]{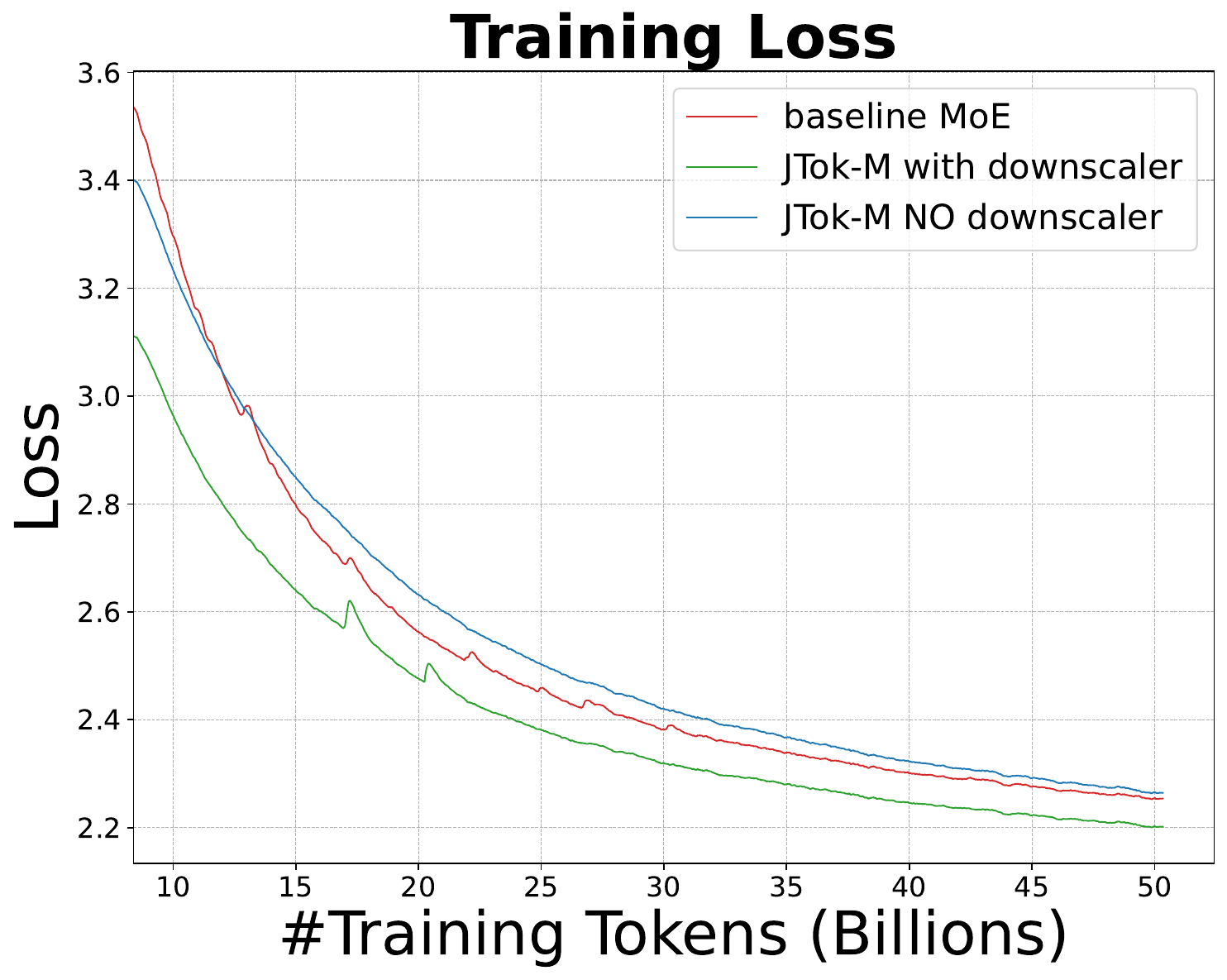}
    \label{fig:scaler_ablation_loss}
\end{subfigure}
\hfill
\begin{subfigure}[b]{0.48\textwidth}
    \centering
    \includegraphics[width=\textwidth]{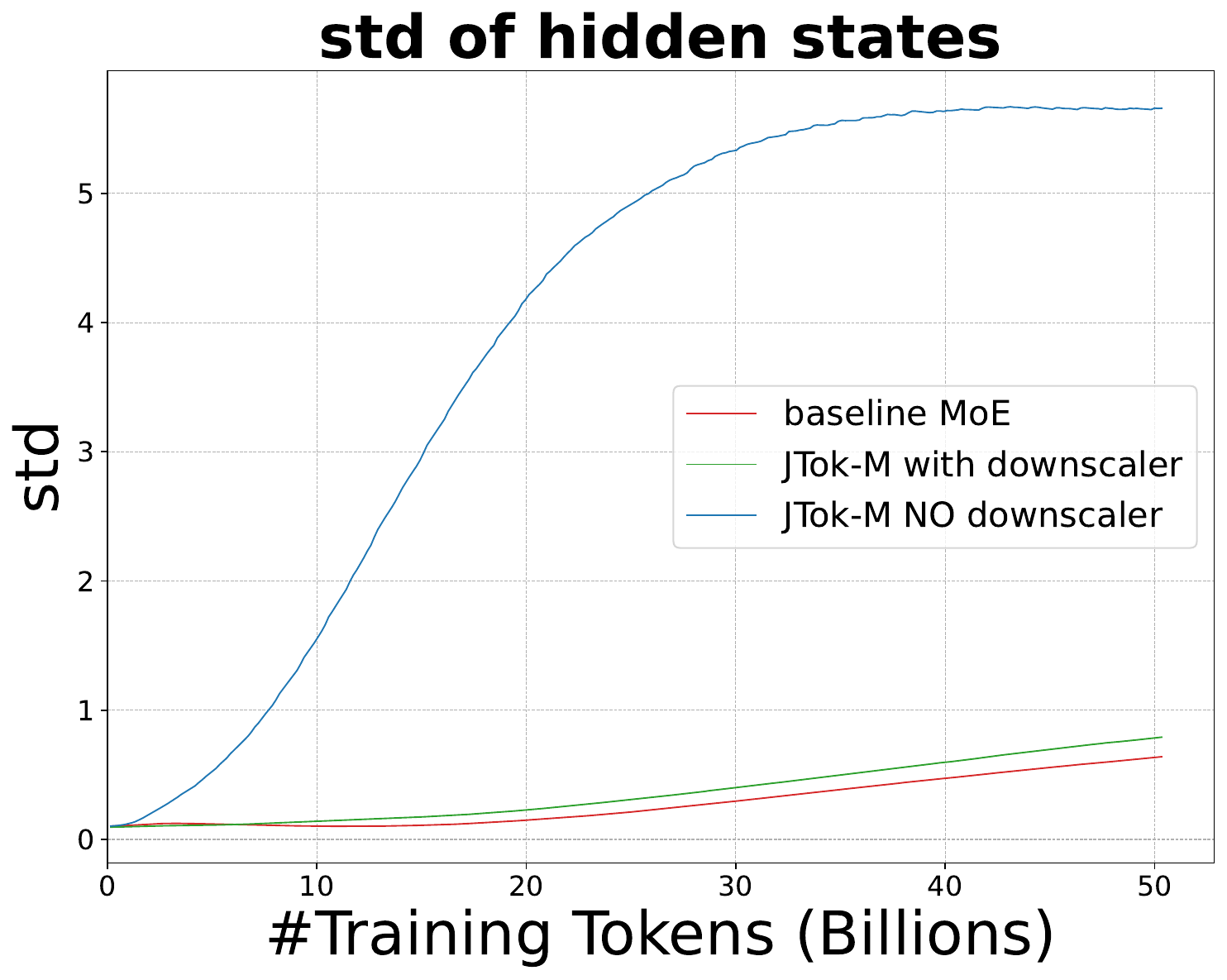}
    \label{fig:scaler_ablation_std}
\end{subfigure}

\vspace{-5pt}
\caption{
Ablation of the scaling factor $\frac{1}{\sqrt{2N_\ell}}$ in JTok-M. Left: training loss. Right: averaged std of hidden states over layers and tokens.}

\label{fig:JTok-M_scale_ablation}
\vspace{-7pt}
\end{figure}

Fig.~\ref{fig:JTok-M_scale_ablation} (left) reports the training loss, while Fig.~\ref{fig:JTok-M_scale_ablation} (right) shows the standard deviation of hidden states averaged over all layers and tokens. Without the scaling factor, the activation std grows rapidly to above $5.5$ within the first 50B training tokens and the model underperforms the baseline MoE in terms of loss. In contrast, JTok-M with the proposed $\frac{1}{\sqrt{2N_\ell}}$ factor keeps variance of hidden states close to baseline in a well-behaved range and consistently achieves lower training loss than the baseline. These results confirm that the scaling factor is crucial to prevent variance explosion and to ensure that JTok-M’s modulation effectively improves optimization.

\subsection{Norm in JTok/JTok-M}
\label{app:ablation_norm}

JTok augments each layer with a token-indexed modulation vector retrieved from an embedding table, as described in Section~\ref{JTok}.

\textbf{Why normalizing the modulation vector matters.}
At a high level, JTok aims to provide \emph{token-specific} and dimension-wise modulation.
In this mechanism, the direction of $\mathbf{E}^{\ell}[x]$ encodes how the residual update should be modulated across hidden dimensions, whereas its magnitude primarily controls the overall strength of that modulation.
Applying $\mathrm{Norm}_{\varepsilon}(\cdot)$ explicitly removes the magnitude degree of freedom and places the modulation vector on a hypersphere.

This has two practical benefits.

\textbf{1. Better-conditioned optimization: learning direction without chasing scale.}
Without normalization, the gate depends on the raw embedding magnitude, 
so training needs to jointly tune both direction and norm of $\mathbf{E}^{\ell}[x]$ to reach an effective modulation regime.
In long-horizon training with Adam-style optimizers, the effective step magnitude in parameter space is typically bounded by the scheduled learning rate (up to the factor $m/\sqrt{v}$), and in common regimes one often has $m/\sqrt{v}\approx \mathcal{O}(1)$~\cite{adam, ngpt}.
Therefore,  as $\mathbf{E}^{\ell}[x]$ is usually initialized at a small scale~\cite{deepseekv2}, its norm tends to remain $\mathcal{o}(1)$ for a long time unless gradients consistently push in the radial direction.
This makes it difficult for the model to quickly reach the ``right'' modulation strength through embedding norms alone.
Normalization resolves this by decoupling direction learning from scale: the token embedding focuses on learning a direction on the hypersphere, while the learnable per-dimension scaler $\mathbf{s}^{\ell}$ controls the overall modulation strength.
Although normalization fixes one scalar degree of freedom, it retains essentially all expressive power in high dimensions. Noting that a $d$-dimensional vector on the hypersphere still has $d-1$ degrees of freedom.

\textbf{2. Stable and comparable modulation across tokens.}
Token-indexed tables can have highly non-uniform access frequencies, which leads to heterogeneous gradient statistics across token embeddings.
Without normalization, different tokens can drift to different norms, making the gate distribution highly inconsistent and potentially harming stability (e.g., some tokens barely modulate while others over-modulate).
By constraining all retrieved vectors to a comparable norm, it produces a more predictable gate scale across tokens and layers, making the plugin behave like a controlled residual modulation.

\begin{figure*}[t]
   \centering
   \begin{subfigure}[b]{0.48\textwidth}
      \includegraphics[width=\textwidth]{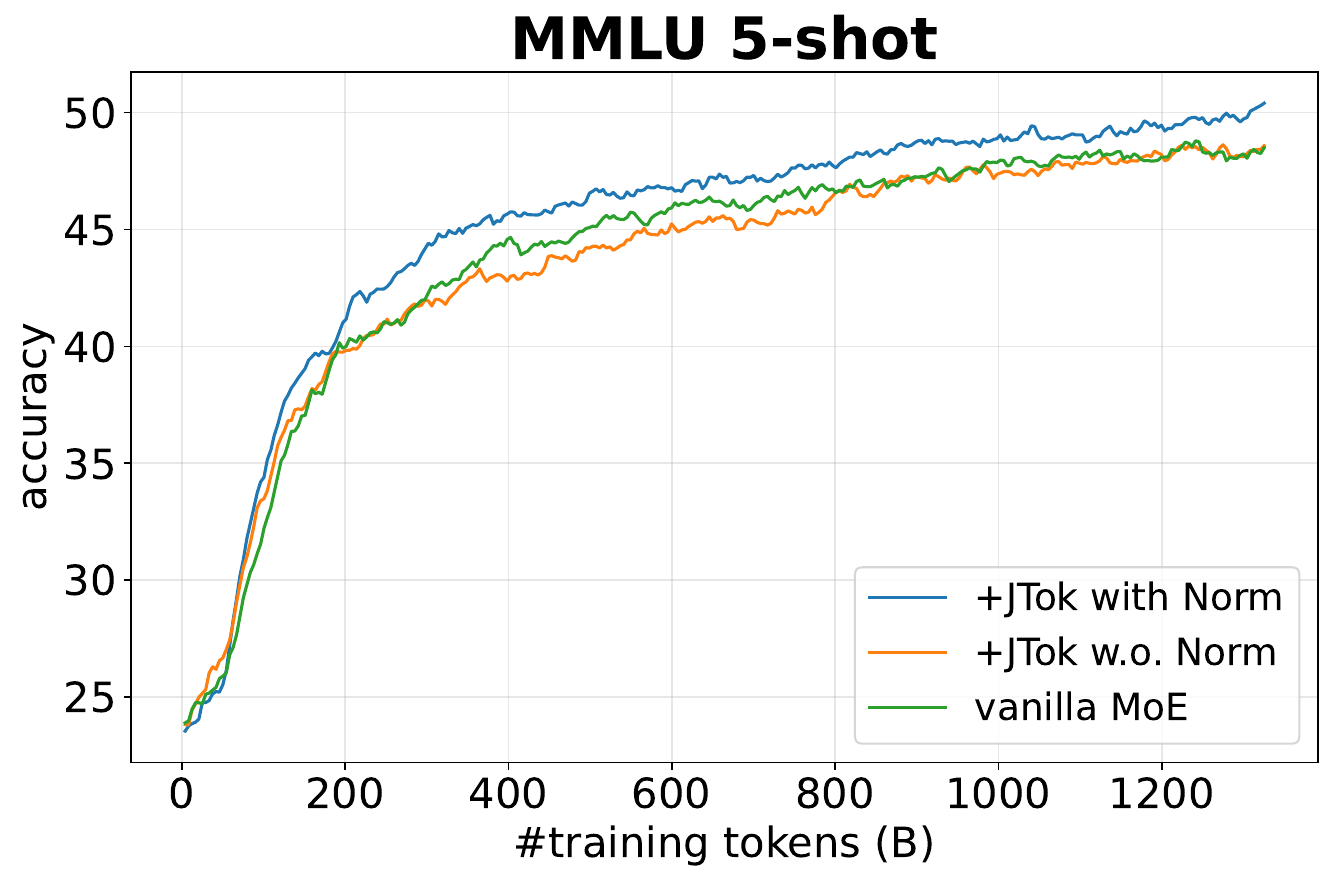}
      \caption{MMLU accuracy}
      \label{fig:ablation_norm_mmlu} 
   \end{subfigure}
   \hfill
   \begin{subfigure}[b]{0.48\textwidth}
      \includegraphics[width=\textwidth]{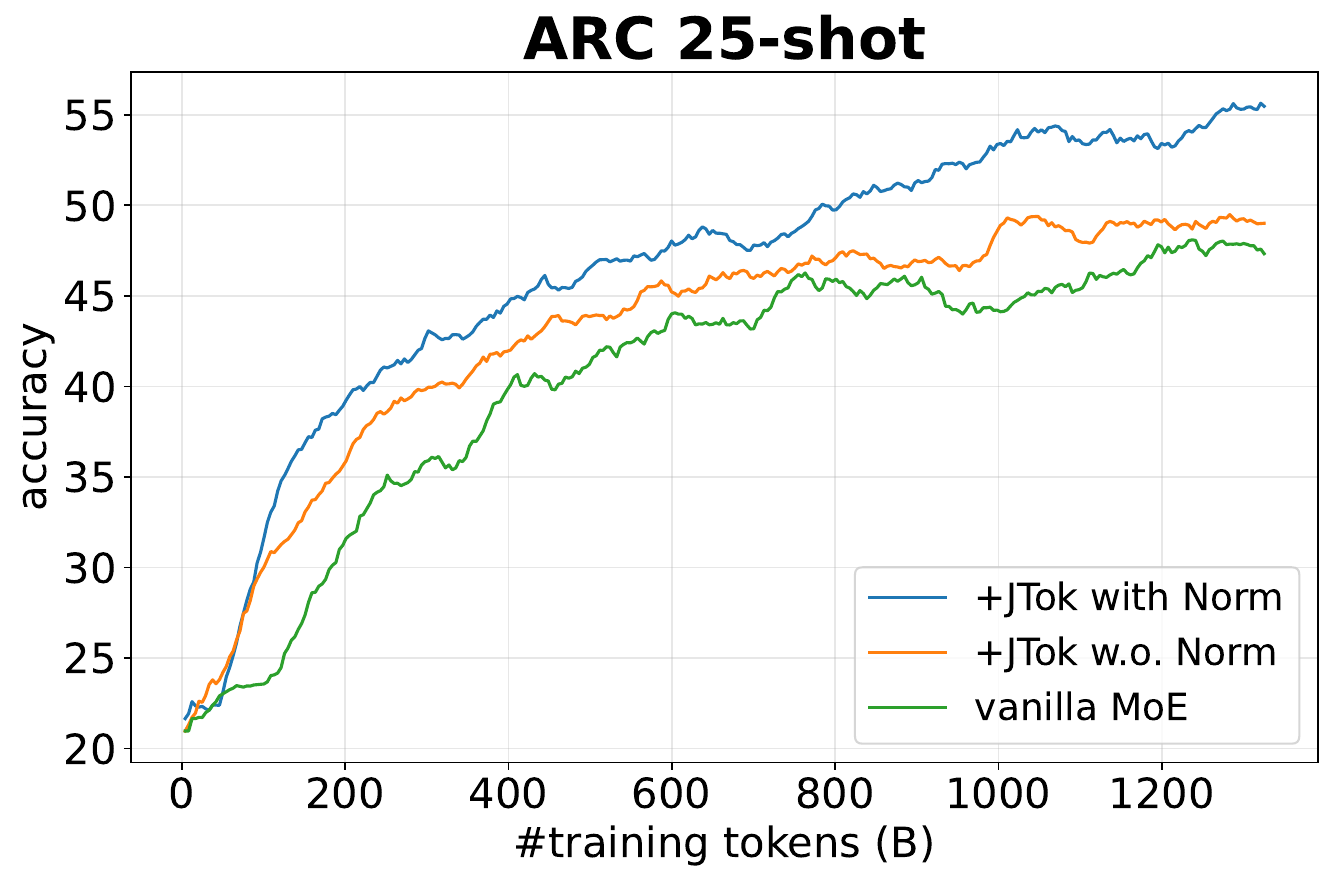}
      \caption{ARC accuracy}
      \label{fig:ablation_norm_arc}
   \end{subfigure}
   
   \vspace{5pt} % Add some vertical space between rows
   
   \begin{subfigure}[b]{0.48\textwidth}
      \includegraphics[width=\textwidth]{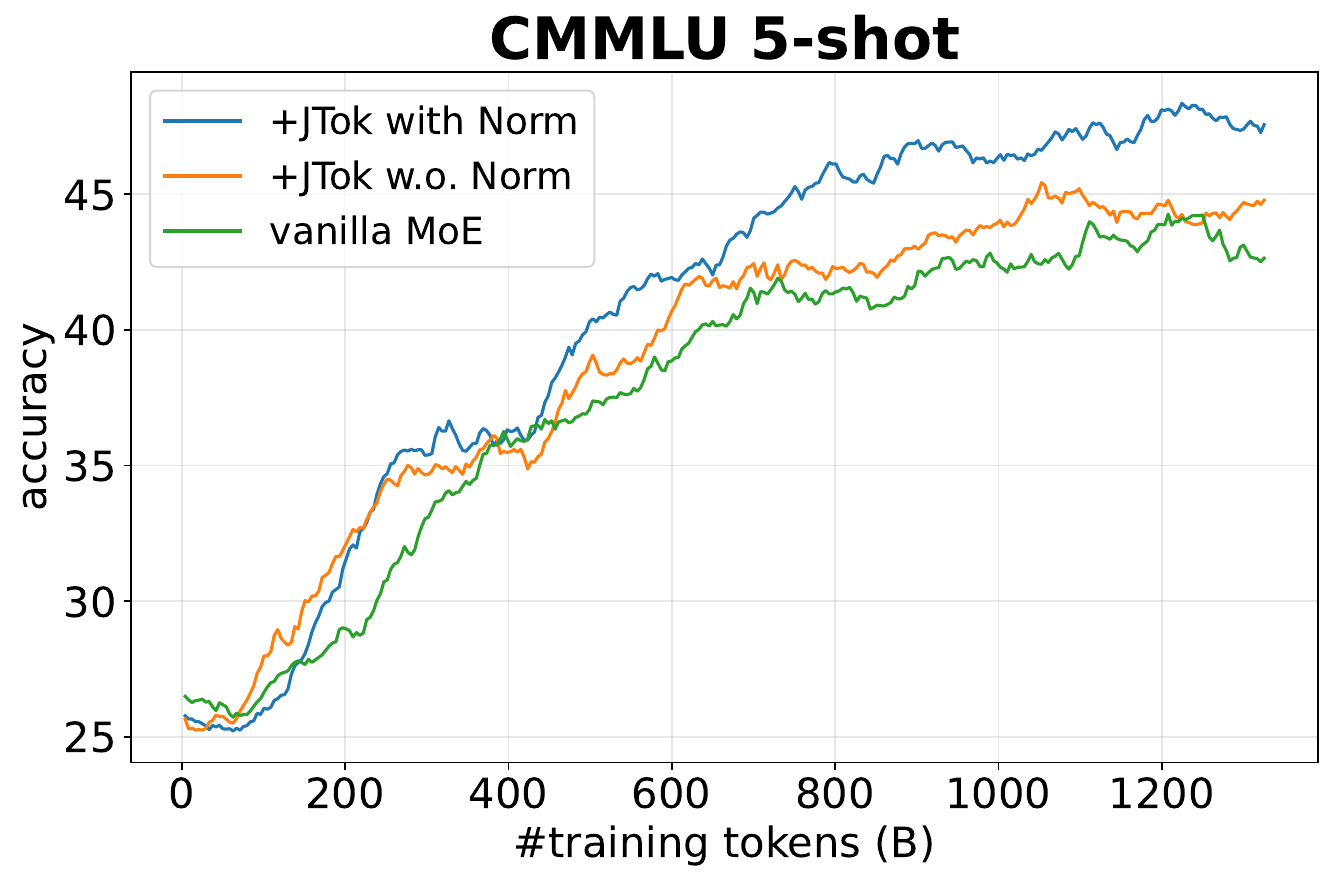}
      \caption{CMMLU accuracy}
      \label{fig:ablation_norm_cmmlu}
   \end{subfigure}
   \hfill
   \begin{subfigure}[b]{0.48\textwidth}
      \includegraphics[width=\textwidth]{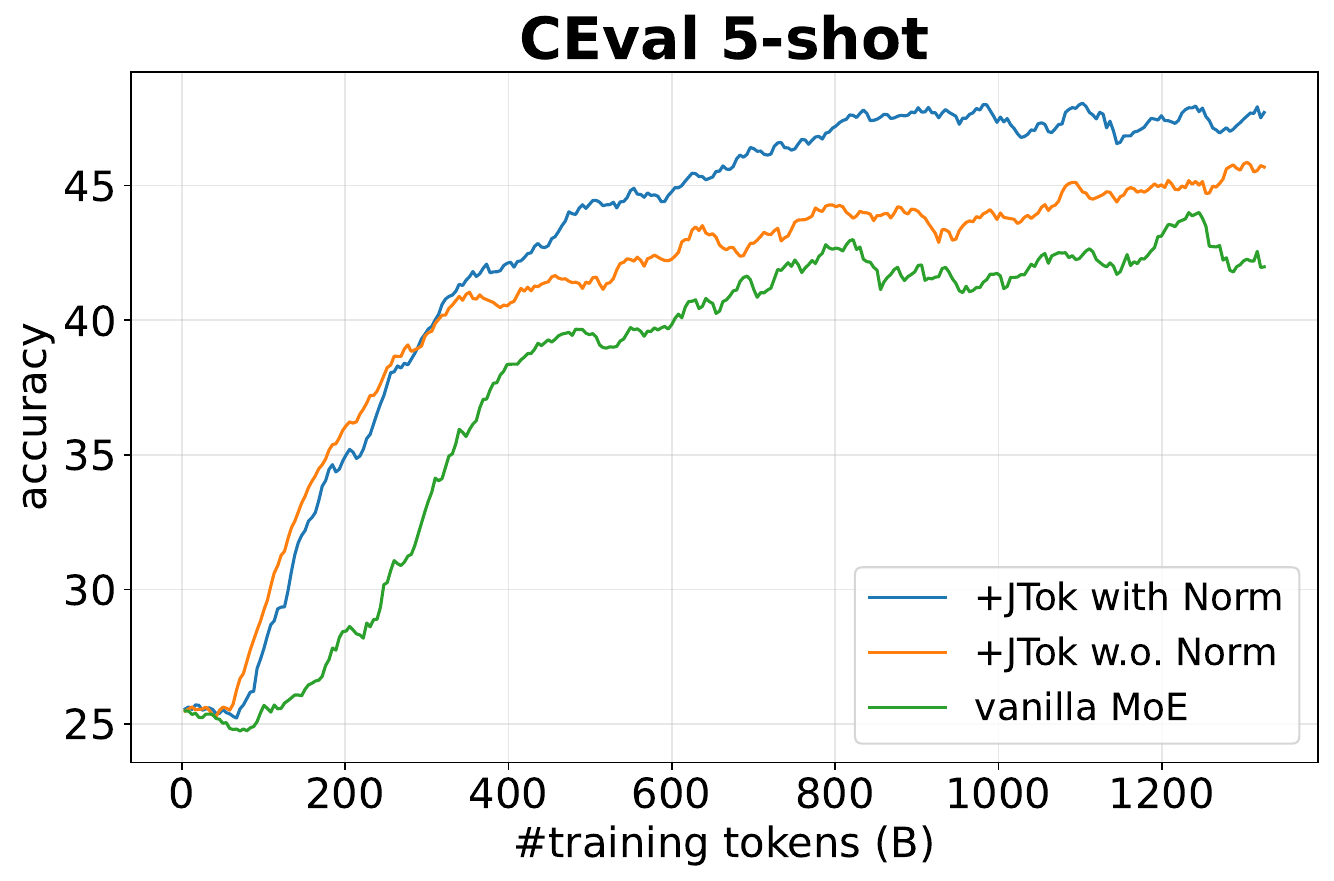} % Placeholder filename
      \caption{CEval accuracy}
      \label{fig:ablation_norm_ceval}
   \end{subfigure}
   
   \vspace{-5pt}
   \caption{\textbf{Normalization ablation for JTok.} Downstream accuracy trajectories on MMLU, ARC, CMMLU, and CEval during long-horizon pretraining ($\sim$1.3T tokens) on the 3.2B-A0.5B MoE backbone, comparing backbone-only, JTok with norm, and JTok without norm. Applying norm consistently accelerates convergence and improves the final accuracy across all benchmarks, indicating that hypersphere-normalized modulation yields more stable optimization and a higher performance ceiling.}
   \label{fig:ablation_norm}
   \vspace{-10pt}
\end{figure*}

\paragraph{Experimental setup.}
We conduct an ablation on the 3.2B-A0.5B MoE backbone, comparing three variants:
(i) backbone-only, (ii) JTok w/ norm using $\mathrm{Norm}_{\varepsilon}$ as above, and (iii) JTok w.o. norm where the retrieved vector is used directly without normalization.
All runs use the same architecture and training recipe; to better expose optimization differences, we intentionally \emph{overtrain} to $\sim$1.3T tokens and periodically evaluate downstream accuracy throughout training.
We report MMLU, ARC, CMMLU, and CEval trajectories in Fig.~\ref{fig:ablation_norm}.

\paragraph{Results and discussion.}
Fig.~\ref{fig:ablation_norm} shows that normalization substantially improves both optimization speed and final performance.
Across all four benchmarks, JTok w/ norm rises faster in the early-to-mid training regime and consistently maintains a higher accuracy curve than JTok w.o. norm.
Moreover, the gap persists (or even widens) even under long training, indicating that normalization is not merely a transient stabilization trick but also enables JTok to realize a higher effective capacity ceiling.

\section{Detailed Training and Model Hyperparameters} \label{app:hyper}
Table~\ref{tab:hyper-pretrain-dense} provides the pretraining hyperparameters for dense models, covering four model sizes (small, medium, large, and XL) and their JTok variants. Table~\ref{tab:hyper-pretrain-moe} shows the pretraining hyperparameters for MoE models, specifically two configurations (1.5B-A250M, 3.2B-A0.5B and 17B-A2B) together with their JTok and JTok-M variants.

\begin{table}[tb!]
    \centering
    \caption{Hyper-parameters for dense model pretraining.}
    \label{tab:hyper-pretrain-dense}
    \vspace{-5pt}
    \resizebox{0.8\textwidth}{!}{
    \begin{tabular}{cccccc}
        \toprule & \textbf{Parameters} & \textbf{small} & \textbf{medium} & \textbf{large} & \textbf{XL} \\
        \midrule
         \multirow{7}{*}{Optimizer} & lr-schedule & cosine & cosine & cosine & WSD~\citep{wsd} \\
         & max, min lr & (1e-3, 1e-4) & (8e-4, 8e-5) & (6e-4, 6e-5) & (6e-4, 0) \\
         & warmup-ratio & \multicolumn{4}{c}{0.05} \\
         & decay-ratio & 0.95 & 0.95 & 0.95 & 0.00 \\
        & optimizer & \multicolumn{4}{c}{AdamW~\cite{adam}} \\
        & weight-decay & \multicolumn{4}{c}{0.1} \\
         & grad\_clip & \multicolumn{4}{c}{1.0}\\
        \midrule
         \multirow{6}{*}{Backbone} & \#params & 190M & 0.5B & 1B & 1.5B \\
         & hidden dim. & 768 & 1024 & 1280 & 1536 \\
        & \#layers & 12 & 24 & 36 & 28 \\
         & \#q heads & 12 & 16 & 20 & 12 \\
         & \#kv heads & 12 & 16 & 20 & 2  \\
         & context-length & 1024 & 1024 & 1024 & 8192 \\
         & FFN size & 3072 & 4096 & 5120 & 8960 \\
        \midrule
        \multirow{2}{*}{Data} & vocab size & 50304 & 50304 & 50304 & 152064 \\
        & \#tokens(B) & 100 & 100 & 100 & 300  \\
         & batch size & 4096 & 4096 & 4096 & 512 \\
        \midrule
         JTok & \#extra params. & 464M & 1.2B & 2.3B & 6.5B \\
        \bottomrule
    \end{tabular}
    }
\vspace{-5pt}
\end{table}

\begin{table}[tb!]
    \centering
    \caption{Hyper-parameters for MoE model pretraining.}
    \label{tab:hyper-pretrain-moe}
    \vspace{-5pt}
    \resizebox{0.6\textwidth}{!}{
    \begin{tabular}{ccccc}
        \toprule & \textbf{Parameters} & \textbf{1.5B-A250M} & \textbf{3.2B-A0.5B} & \textbf{17B-A2B} \\
        \midrule
         \multirow{7}{*}{Optimizer} & lr-schedule & \multicolumn{3}{c}{WSD} \\
         & lr & \multicolumn{3}{c}{4.2e-4} \\
         & warmup-ratio & \multicolumn{3}{c}{0.05} \\
         & decay-ratio & \multicolumn{3}{c}{0.00} \\
        & optimizer & \multicolumn{3}{c}{AdamW} \\
        & weight-decay & \multicolumn{3}{c}{0.1} \\
         & grad\_clip & \multicolumn{3}{c}{1.0} \\
        \midrule
         \multirow{11}{*}{Backbone} & hidden dim. & 512 & 768 & 2048 \\
         & \#dense layers & 1 & 1 & 1 \\
         & \#moe layer & 11 & 17 & 27 \\
         & \#q heads & 8 & 16 & 16 \\
         & \#kv heads & 4 & 4 & 8 \\
         & context-length & 8192 & 8192 & 8192 \\
         & \#routed experts & 144 & 144 & 64\\
         & \# shared experts & \multicolumn{3}{c}{1} \\
         & topK route & 8 & 8 & 6 \\
         & dense FFN size & \multicolumn{3}{c}{10944} \\
         & MoE FFN size & 512 & 512 & 1408 \\
        \midrule
        \multirow{2}{*}{Data} & vocab size & \multicolumn{3}{c}{152064} \\
        & batch size & 1024 & 1024 & 2048 \\
        & \#tokens(B) & 300 & 500 & 570 \\
       \midrule
        JTok & \#extra params. & 934M & 2.1B & - \\
       \midrule
        \multirow{3}{*}{JTok-M} & $n_e$ & \multicolumn{3}{c}{5} \\
        & K & \multicolumn{3}{c}{2} \\
        & \#extra params. & 4.7B & 10.5B & 43.6B \\
        \bottomrule
    \end{tabular}
    }
\vspace{-10pt}
\end{table}

\end{document}